\documentclass[twoside,11pt]{article}
\usepackage{journal2e_v1}

\usepackage{epsfig}
\usepackage{epsf}
\usepackage{multicol}
\usepackage{subfigure}
\usepackage{amsmath}
\usepackage{amssymb}
\usepackage{latexsym}
\usepackage{algorithm}
\usepackage{algorithmic}

\def\b{{\bf b}}

\def\K{{\bf K}}
\def\k{{\bf k}}

\def\x{{\bf x}}

\def\0{{\bf 0}}
\def\1{{\bf 1}}

\def\CM{{\mathcal C}}
\def\FM{{\mathcal F}}

\def\TM{{\mathcal T}}

\def\XM{{\mathcal X}}
\def\YM{{\mathcal Y}}

\def\RB{{\mathbb R}}

\def\EB{{\mathbb E}}

\def\bet{\mbox{\boldmath$\beta$\unboldmath}}

\def\sign{\mathrm{sign}}

\def\argmin{\mathop{\rm argmin}}

\title{Coherence Functions with Applications in Large-Margin Classification Methods}

\author{\name Zhihua Zhang and Guang Dai \\
\addr College of Computer Science \& Technology \\
Zhejiang University \\
Hangzhou, China 310027 \\
\texttt{\{zhzhang, guang.gdai\}@cs.zju.edu.cn} \AND
\name Michael I. Jordan\\
\addr  Computer Science Division and Department of Statistics \\
University of California, Berkeley \\
CA 94720-1776 USA \\
\texttt{jordan@eecs.berkeley.edu}
}

\date{Revised, February  22, 2012}

\begin{document}

\maketitle

\begin{abstract}
Support vector machines (SVMs) naturally embody sparseness due to 
their use of hinge loss functions. However, SVMs can not directly estimate
conditional class probabilities. In this paper we propose and study a family of
\emph{coherence functions}, which are convex and differentiable, as 
surrogates of the hinge function. The coherence function is derived 
by using the maximum-entropy principle and is characterized by a temperature 
parameter. It bridges the hinge function and the logit function in logistic 
regression.  The limit of the coherence function at zero temperature 
corresponds to the hinge function, and the limit of the minimizer of 
its expected error is the minimizer of  the expected error of the hinge 
loss.  We refer to the use of the coherence function in large-margin 
classification as ``\emph{${\cal C}$-learning},'' and we present efficient
coordinate descent algorithms for the training of regularized ${\cal C}$-learning 
models.
\end{abstract}

%%%%%%%%%%%%%%%%%%%%%%%%%%%%%%%%%%%%%%%%%%%%%%%%%%%%%%%%%%%%%%%%%%%%%%%%%%
%%%%%%%%%%%%%%%%%%%%%%%%%%%%%%%%%%%%%%%%%%%%%%%%%%%%%%%%%%%%%%%%%%%%%%%%%%
\section{Introduction}

Large-margin classification methods have become increasingly popular
since the advent of boosting~\citep{FreundIC:1995}, support vector machines
(SVM)~\citep{Vapnik:1998} and their variants such as $\psi$-learning~\citep{ShenJASA:2003}.
Large-margin classification methods are typically devised based on
a majorization-minimization procedure, which approximately solves an otherwise intractable optimization problem defined with the 0-1 loss.
For example, the conventional SVM employs a hinge loss, the AdaBoost algorithm employs the exponential loss, and $\psi$-learning employs a
so-called $\psi$-loss, as majorizations of the 0-1 loss.

Large-margin classification methods can be unified using the tools of
regularization theory;
that is, they can be expressed as the form of ``loss" + ``penalty"~\citep{HastieBook:SL}.  Sparseness
has also emerged as a significant theme generally associated with large-margin methods. Typical approaches for achieving sparseness
are to use either a non-differentiable penalty or a non-differentiable loss.
Recent developments in the former vein focus on the use of a
$\ell_1$ penalty~\citep{TibshiraniLASSO:1996} or the elastic-net penalty
(a mixture of the $\ell_1$ and $\ell_2$ penalties)~\citep{ZouENET:2005}
instead of the  $\ell_2$ penalty which is typically used in
large-margin classification methods.  As for non-differentiable
losses, the paradigm case is the hinge loss function that is used
for the SVM and which leads to a sparse expansion of the discriminant
function.

Unfortunately, the conventional SVM does not directly estimate a conditional
class probability. Thus, the conventional SVM is unable to provide estimates
of uncertainty in its predictions---an important desideratum in real-world
applications.
Moreover, the non-differentiability of the hinge loss also makes it difficult to extend  the conventional SVM to multi-class
classification problems. Thus,  one seemingly natural approach to constructing  a classifier
for the binary and multi-class problems is to consider a smooth loss function,
while an appropriate penalty is employed to maintain the sparseness of the classifier.
For example, regularized logistic regression
models based on logit losses~\citep{FriedmanHastieTibshirani08:2008} are competitive with SVMs.

Of crucial concern are the statistical properties~\citep{LinSVM:2002,BartlettJordanMcauliffe:2006,ZhangAS:2004} of the majorization
function for the original 0-1 loss function.
In particular, we analyze the statistical properties of extant
majorization functions, which are built on the exponential, logit
and hinge functions. This analysis inspires us to propose a new
majorization function, which we call a \emph{coherence function} due
to a connection with statistical mechanics. We also define a loss
function that we refere to as $\CM$-loss based on the coherence function.

The $\CM$-loss is smooth
and convex, and it
satisfies a Fisher-consistency condition---a desirable statistical property~\citep{BartlettJordanMcauliffe:2006,ZhangAS:2004}.
The $\CM$-loss has
the advantage over the hinge loss  that it provides an estimate of the
conditional class probability, and over the logit loss  that
one limiting version of it is just the hinge loss. 
Thus, the $\CM$-loss as well as the coherence function have
several desriable properties in the context of large-margin classifiers.

In this paper we show how the coherence function can be used to develop
an effective approach to estimating the class probability of the conventional 
binary SVM.  \citet{Platt:1999} first exploited a sigmoid link function 
to map the SVM outputs into probabilities, while \citet{SollichML:2002}
used logarithmic scoring rules~\citep{Bernardo:1994} to transform the hinge loss
into the negative of a conditional log-likelihood (i.e., a predictive class probability).
Recently, \citet{WangShen:2008} developed an interval estimation method.
Theoretically, \citet{SteinwartJMLR:2003}
and \citet{BartlettJMLR:2007} showed that the class probability can be
asymptotically estimated by replacing the hinge loss with a differentiable
loss. Our approach also appeals to asymptotics to derive a method
for estimating the class probability of the conventional binary SVM.

Using the $\CM$-loss, we devise  new large-margin classifiers which
we refer to as \emph{$\CM$-learning}.  To maintain sparseness, we
use the elastic-net penalty in addition to $\CM$-learning.
We in particular propose two versions. The first version is based
on reproducing kernel Hilbert spaces (RKHSs) and it can automatically
select the number of support vectors via penalization.  The second
version focuses on the selection of features again via penalization.
The classifiers  are trained by  coordinate descent algorithms developed
by \citet{FriedmanHastieTibshirani08:2008} for generalized linear models.

The rest of this paper is organized as follows.  In Section~\ref{sec:form} we
summarize the fundamental basis of large-margin classification.
Section~\ref{sec:c_loss} presents $\CM$-loss functions,  their mathematical
properties, and a method for class probability
estimation of the conventional SVM.
Section~\ref{sec:c_learn} studies our $\CM$-learning algorithms.
We conduct an experimental analysis
in Section~\ref{sec:examp} and conclude our work in
Section~\ref{sec:concl}. All proofs are deferred to the Appendix.

%%%%%%%%%%%%%%%%%%%%%%%%%%%%%%%%%%%%%%%%%%%%%%%%%%%%%%%%%%%%%%%%%%%%%
%%%%%%%%%%%%%%%%%%%%%%%%%%%%%%%%%%%%%%%%%%%%%%%%%%%%%%%%%%%%%%%%%%%%%
\section{Large-Margin Classifiers} \label{sec:form}

We consider a \emph{binary} classification
problem with a set of training data $\TM=\{\x_i, y_i\}_1^{n}$, where
$\x_i \in \XM \subset \RB^d$ is an input vector and $y_i \in \YM = \{1, -1\}$
is the corresponding class label. 
Our goal is to find a decision function $f(\x)$ over a measurable function class $\FM$.
Once such an $f(\x)$ is obtained,
the classification rule is
$y = \sign(f(\x))$ where $\sign(a)=1, 0, -1$ according to $a>0$, $a=0$ or $a<0$. Thus, we have that $\x$ is misclassified if and
only if $y f(\x)\leq 0$  (here we ignore the case that $f(\x)=0$).

Let $\eta(\x)=\mathrm{Pr}(Y=1|X=\x)$ be the conditional probability
of class 1 given $\x$ and  let $P(X, Y)$ be the probability distribution over $\XM \times \YM$. For a measurable decision function $f(\x): \XM \rightarrow \RB$,
the expected error at $\x$ is then defined
by
%\begin{equation} \label{eq:exp_error}
\[
\Psi(f(\x)) =\EB(I(Y f(X) \leq 0)|X=\x) = I_{[f(\x) \leq 0]} \eta(\x) + I_{[f(\x) > 0]}(1-\eta(\x)),
\]
%\end{equation}
where $I_{[\#]}=1$ if $\#$ is true and 0 otherwise. The generalization error is
\[
\Psi_{f} = \EB_{P}  I_{[Y f(X)\leq 0]}=\EB_{X}\big[I_{[f(X) \leq 0]} \eta(X) + I_{[f(X) > 0]}(1-\eta(X))\big],
\]
where the expectation $\EB_{P}$ is taken with respect to the distribution $P(X, Y)$ and $\EB_{X}$ denotes the expectation over the input data $\XM$.
The optimal Bayes error is $\hat{\Psi}=\EB_{P} I_{[Y(2\eta(X){-}1)\leq 0]}$, which is the minimum  of $\Psi_{f}$ with respect to measurable functions $f$.

A classifier is a classification algorithm which finds a measurable function $f_{\TM}: \XM \rightarrow \RB$ based on the training data $\TM$.
We assume that the $(\x_i, y_i)$ in $\TM$ are independent and identically distributed from $P(X, Y)$.
A classifier is said to be \emph{universally consistent} if
\[
\lim_{n \rightarrow \infty} \Psi_{f_{\TM}} = \hat{\Psi}
\]
holds in probability for any distribution $P$ on $\XM{\times}\YM$.  It is strongly universally consistent
if the condition $\lim_{n \rightarrow \infty} \Psi_{f_{\TM}} = \hat{\Psi}$
is satisfied almost surely~\citep{SteinwartTIT:2005}.

The empirical generalization error on the training data $\TM$ is given by
%\begin{equation} \label{eq:emp_error}
\[
\Psi_{emp}=\frac{1}{n} \sum_{i=1}^n I_{[y_i f(\x_i)\leq 0]}.
\]
%\end{equation}
Given that the empirical generalization error $\Psi_{emp}$ is equal to its minimum value zero when all
training data are correctly classified, we wish to use $\Psi_{emp}$ as
a basis for devising classification algorithms. However, the corresponding minimization problem is computationally intractable.

%%%%%%%%%%%%%%%%%%%%%%%%%%%%%%%%%%%%%%%%%%%%%%%%%%%%%%%%%
\subsection{Surrogate Losses} \label{sec:surrogate}

A wide variety of classifiers
can be understood as minimizers of a continuous \emph{surrogate loss} function $\phi(y f(\x))$,
which upper bounds the 0-1 loss $I_{[y f(\x)\leq 0]}$. Corresponding to $\Psi(f(\x))$ and $\Psi_{f}$, we denote
$R(f(\x)) = \phi(f(\x)) \eta(\x) + \phi(-f(\x))(1-\eta(\x))$ and
\[
R_{f} = \EB_{P} [\phi(Y f(X))] = \EB_{X}\big[ \phi(f(X)) \eta(X) + \phi(-f(X))(1-\eta(X))\big].
\]
For convenience, we assume that $\eta \in [0, 1]$ and define the notation
\[
R(\eta, f) = \eta \phi(f) + (1-\eta) \phi(-f).
\]
The surrogate $\phi$ is said to be Fisher consistent, if for every $\eta \in [0, 1]$ the minimizer of $R(\eta, f)$
with respect to $f$ exists and is unique and the minimizer (denoted $\hat{f}(\eta)$) satisfies
$\sign(\hat{f}(\eta))= \sign(\eta -1/2)$
\citep{LinSVM:2002,BartlettJordanMcauliffe:2006,ZhangAS:2004}.
Since $\sign(u)=0$ is equivalent to $u=0$, we have that $\hat{f}(1/2)=0$.
Substituting  $\hat{f}(\eta)$ into $R(\eta, f)$, we also define the following notation:
\[
\hat{R}(\eta) = \inf_{f} R(\eta, f) = R(\eta, \hat{f}(\eta)).
\]
The difference between $R(\eta, f)$ and $\hat{R}(\eta)$ is
\[
\triangle R(\eta, f) = R(\eta, f) - \hat{R}(\eta) = R (\eta, f) - R (\eta, \hat{f}(\eta)).
\]

When regarding $f(\x)$ and $\eta(\x)$ as functions of $\x$, it is clear that $\hat{f} (\eta(\x))$
is the minimizer of $R (f(\x))$ among all measurable function class $\FM$. That is,
\[
\hat{f} (\eta(\x)) = \argmin_{f(\x) \in \FM} R (f(\x)).
\]
In this setting, the difference between $R_{f}$ and  $\EB_{X} [R (\hat{f} (\eta(X)))]$ (denoted ${R}_{\hat{f}}$) is given by
\[
\triangle R_{f} = R_{f} - {R}_{\hat{f}} = \EB_{X} \triangle R (\eta(X), f(X)).
\]

If $\hat{f}(\eta)$ is invertible,  then the inverse function $\hat{f}^{-1}(f(\x))$ over $\FM$ can be regarded as
a class-conditional probability estimate given that $\eta(\x)=\hat{f}^{-1}(\hat{f}(\x))$. Moreover, \citet{ZhangAS:2004} showed that
$\triangle R_{f}$ is the expected distance between the conditional probability $\hat{f}^{-1}(f(\x))$ and the true
conditional probability $\eta(\x)$. Thus, minimizing $R_f $ is equivalent to minimizing the
expected distance between  $\hat{f}^{-1}(f(\x))$ and  $\eta(\x)$.

\begin{table*}[!tb]
\begin{center}
\caption{Surrogate losses for margin-based classifier.} \label{tab:binary_class}
\begin{tabular}{c|c|c|c} \hline
             {\tt Exponential Loss}     & {\tt Logit Loss}  & {\tt Hinge Loss} & {\tt Squared Hinge Loss} \\
\hline
 $\exp[{-} y f(\x)/2]$  &
$\log[1{+} \exp({-} y
f(\x))]$  & $[1-y f(\x)]_{+}$ & $([1-y f(\x)]_{+})^2$ \\
\hline
\end{tabular}
\end{center}
\end{table*}

Table~\ref{tab:binary_class} lists four common surrogate functions used in
large-margin classifiers. Here
$[u]_{+}=\max\{u, 0\}$ is a so-called hinge function and $([u]_{+})^2=(\max\{u, 0\})^2$ is a squared hinge function
which is used for developing the $\ell_2$-SVM~\citep{CristianiniBook:2000}. Note that
we typically scale the logit loss to equal $1$ at $y f(\x)=0$. These functions are
convex and the upper bounds of $I_{[y f(\x)\leq 0]}$.
Moreover, they are Fisher consistent. In particular, the following
result has been established by \citet{FriedmanAS:2000} and \citet{LinSVM:2002}.
\begin{proposition} \label{prop:1} Assume that $0<\eta(\x) <1$ and $\eta(\x)\neq 1/2$. Then, the minimizers
of $\EB(\exp[- Y f(X)/2]|X=\x)$ and $\EB(\log[1+\exp(- Y f(X)) ]|X =\x)$ are both $\hat{f}(\x)  = \log \frac{\eta(\x)}{1-\eta(\x)}$,
the minimizer of $\EB\big([1- Y f(X)]_{+}|X=\x\big)$ is $\hat{f}{(\x)} = \sign(\eta(\x)-1/2)$, and the minimizer
of $\EB\big(([1- Y f(X)]_{+})^2|X=\x\big)$ is $\hat{f}{(\x)} = 2 \eta(\x)-1$.
\end{proposition}

When the exponential or logit loss function is used, $\hat{f}^{-1}(f(\x))$ exists. It is clear that $\eta(\x) = \hat{f}^{-1}(\hat{f}(\x))$.  For any $f(\x) \in \FM$, we denote  the inverse function by $\tilde{\eta}(\x)$, which
is
%\begin{equation} \label{eq:logit_m}
\[
\tilde{\eta}(\x) =  \hat{f}^{-1}(f(\x)) = \frac{1}{1 + \exp(-f(\x))}.
\] %\end{equation}
Unfortunately, the minimization of the hinge loss (which is the basis of
the SVM) does not yield a class probability estimate~\citep{LinSVMML:2002}.

\subsection{The Regularization Approach}

Given a surrogate loss function $\phi$, a large-margin classifier typically
solves the following optimization problem:
\begin{equation} \label{eq:margin_m}
\min_{f \in \FM} \frac{1}{n} \sum_{i=1}^n \phi(y_i f(\x_i)) + \gamma J(h),
\end{equation}
where $f(\x) = \alpha + h(\x)$, $J(h)$ is a regularization term to penalize model complexity and $\gamma$ is the degree of penalization.

Suppose that  $f = \alpha + h \in (\{1\} + {\cal H}_{K})$ where ${\cal H}_K$ is a reproducing kernel Hilbert space (RKHS) induced by a reproducing kernel
$K(\cdot, \cdot): \XM{\times}\XM \rightarrow \RB$.
Finding $f(\x)$ is then formulated as a regularization problem of
the form
\begin{equation} \label{eq:regu}
\min_{f \in {\cal H}_K} \left\{ \frac{1}{n} \sum_{i=1}^{n} \phi(y_if(\x_i)) + \frac{\gamma}{2} \| h\|^2_{{\cal H}_K} \right\},
\end{equation}
where $\|h\|^2_{{\cal H}_K}$ is the RKHS norm.
By the representer theorem \citep{Wahba:1990}, the solution of
(\ref{eq:regu}) is of the form
\begin{equation} \label{eq:rth}
f(\x_i) = \alpha + \sum_{j=1}^{n} \beta_j K(\x_i, \x_j) = \alpha + \k_i'
\bet,
\end{equation}
where $\bet=(\beta_1,
\ldots, \beta_n)'$ and $\k_i=(K(\x_i, \x_1), \ldots, K(\x_i, \x_n))'$. Noticing that $\|
h\|^2_{{\cal H}_{K}}= \sum_{i, j=1}^n K(\x_i, \x_j) \beta_i \beta_j$ and
substituting (\ref{eq:rth}) into (\ref{eq:regu}), we obtain the
minimization problem with respect to  $\alpha$ and $\bet$ as
%\begin{equation} \label{eq:svm}
\[
\min_{\alpha, \bet} \bigg\{ \frac{1}{n} \sum_{i=1}^{n} \phi(y_i (\alpha + \k_i'
\bet)) + \frac{\gamma}{2} \bet' \K \bet \bigg\},
\] %\end{equation}
where  $\K=[\k_1,
\ldots, \k_n]$ is the $n{\times}n$ kernel matrix.
Since $\K$ is symmetric and positive semidefinite, the term $\bet' \K \bet$ is in fact an empirical RKHS norm on the training data.

In particular, the conventional SVM defines the surrogate  $\phi(\cdot)$ as the hinge loss and solves the following optimization problem:

\begin{equation} \label{eq:h_svm0}
\min_{f \in \FM} \frac{1}{n} \sum_{i=1}^n [1-y_i (\alpha {+} \k_i'
\bet)]_{+} + \frac{\gamma}{2} \bet'\K \bet.
\end{equation}

In this paper, we are especially interested in \emph{universal kernels}, namely, kernels whose induced RKHS is dense
in the space of continuous functions over $\XM$~\citep{SteinwartJMLR:2001}.   The Gaussian RBF kernel
is such an example.

\subsection{Methods for Class Probability Estimation of SVMs}

Let $\hat{f}(\x)$ be the solution of the SVM problem in (\ref{eq:h_svm0}).
In an attempt to address the problem of class probability estimation for
 SVMs,
\citet{SollichML:2002} proposed a class probability estimate
\[
\hat{\eta}(\x)= \left\{\begin{array}{ll} \frac{1}{1+\exp({-}2
\hat{f}(\x)) } &
\mbox{if } |\hat{f}(\x)|\leq 1, \\
\frac{1}{1+\exp[{-}(\hat{f}(\x){+} \sign(\hat{f}(\x)))]} &
\mbox{otherwise}. \end{array} \right.
\]
This class probability was also
used in the derivation of a so-called complete SVM  by
\citet{MallickJRSSB:2005}.

Another proposal for obtaining class probabilities from SVM
outputs was developed by \citet{Platt:1999}, who employed a
post-processing procedure based on the parametric
formula
%\begin{equation} \label{eq:platt_m}
\[
\hat{\eta}(\x) =  \frac{1}{1 + \exp(A \hat{f}(\x) +B)},
\] %\end{equation}
where the parameters $A$ and $B$ are estimated via the
minimization of the empirical cross-entropy error over the training dataset.

\citet{WangShen:2008} proposed a nonparametric form obtained from
training a sequence of weighted classifiers:
\begin{equation} \label{eq:wsvm}
\min_{f \in \FM} \frac{1}{n} \bigg\{(1-\pi_j) \sum_{y_i=1} [1- y_i f(\x_i)]_{+}   + \pi_j \sum_{y_i=-1} [1-y_i f(\x_i)]_{+}\bigg\} + \gamma J(h)
\end{equation}
for $j=1, \ldots, m{+}1$ such that $0=\pi_1< \cdots < \pi_{m{+}1}=1$. Let $\hat{f}_{\pi_j}(\x)$ be the solution
of (\ref{eq:wsvm}). The estimated class probability is then
$\hat{\eta}(\x) = \frac{1}{2}(\pi_{*}+\pi^{*})$ where $\pi_{*} = \min\{\pi_j: \sign(\hat{f}_{\pi_j}(\x))=-1\}$ and
$\pi^{*} = \max\{\pi_j: \sign(\hat{f}_{\pi_j}(\x))=1\}$.

Additional contributions are due to \citet{SteinwartJMLR:2003}
and \citet{BartlettJMLR:2007}. These  authors showed that the class
probability can be asymptotically estimated by replacing the hinge
loss with various differentiable losses.

%%%%%%%%%%%%%%%%%%%%%%%%%%%%%%%%%%%%%%%%%%%%%%%%%%%%%%%%%%%%%%%%%%%%%
%%%%%%%%%%%%%%%%%%%%%%%%%%%%%%%%%%%%%%%%%%%%%%%%%%%%%%%%%%%%%%%%%%%%%
\section{Coherence Functions}
\label{sec:c_loss}

In this
section we present a smooth and Fisher-consistent majorization
loss, which bridges the hinge loss and the logit loss.  We will see
that one limit of this loss is equal to the hinge loss.
Thus, it is applicable to  the asymptotical estimate of  the class
probability for the conventional SVM  as well as the construction
of margin-based classifiers, which will be presented in
Section~\ref{sec:cpe} and Section~\ref{sec:c_learn}.

%%%%%%%%%%%%%%%%%%%%%%%%%%%%%%%%%%%%%%%%%%%%%%%%%%%%%%%%%%%%%%%%%%%%%
%%%%%%%%%%%%%%%%%%%%%%%%%%%%%%%%%%%%%%%%%%%%%%%%%%%%%%%%%%%%%%%%%%%%%
\subsection{Definition} \label{sec:cohe_f}

Under the $0{-}1$ loss the misclassification costs are specified to
be one, but it is natural to set the misclassification costs to be
a positive constant $u>0$. The empirical generalization error on the
training data is given in this case by
\[
\frac{1}{n} \sum_{i=1}^n u I_{[y_i f(\x_i)\leq 0]},
\]
where $u>0$ is a constant that represents the misclassification
cost. In this setting we can extend the hinge loss as
\begin{equation} \label{eq::h_u}
H_u(y f(\x)) = [u{-}y f(\x)]_{+}.
\end{equation}
It is clear that $H_u(y f(\x)) \geq u I_{[y f(\x)\leq 0]}$. This implies that
$H_u(y f(\x))$ is a majorization of $u I_{[y f(\x)\leq 0]}$.

We apply the maximum entropy principle to develop a smooth surrogate of the hinge loss $[u{-}z]_{+}$. In particular,
noting that $[u{-}z]_{+}=\max\{u{-}z, \; 0\}$, we maximize $w (u{-}z)$ with respect to $w \in (0, 1)$ under
the entropy constraint; that is,
\[
\max_{w \in (0, 1)} \Big\{F = w (u{-} z)
 - {\rho} \big[w \log w + (1-w) \log (1-w) \big]\Big\},
\]
where $-[w\log w + (1{-}w) \log(1{-}w)]$ is the entropy and ${\rho}>0$, a Lagrange
multiplier, plays the role of temperature in thermodynamics.

The maximum of $F$ is
\begin{equation} \label{eq:b_c}
V_{{\rho}, u}(z)= {\rho} \log\big[ 1{+} \exp \frac{u {-}
z} {{\rho}} \big]
\end{equation}
at $w= \exp((u{-}z)/{\rho})/[1+ \exp((u{-}z)/{\rho})]$.
We refer to functions of this form as \emph{coherence functions}
because their properties (detailed in the next subsection) are similar to statistical
mechanical properties  of deterministic annealing~\citep{rose:stat}.

We also consider a scaled variant of $V_{{\rho}, u}(z)$:
\begin{equation} \label{eq:b_closs}
C_{{\rho}, u}(z) = \frac{u}{\log[ 1{+} \exp(u/{\rho})] }\log\big[ 1{+} \exp \frac{u {-}
z} {{\rho}} \big], \; {\rho}>0, \; u > 0,
\end{equation}
which has the property that $C_{{\rho}, u}(z)=u$ when $z=0$.
Recall that $u$ as a misclassification cost should be specified as a positive value. However,
both $C_{{\rho}, 0}(z)$ and $V_{{\rho}, 0}(z)$ are well defined mathematically. Since $C_{{\rho}, 0}(z)=0$ is a trivial case,
we always assume that $u>0$ for $C_{{\rho}, u}(z)$ here and later.
In the binary classification problem,  $z$ is defined as $y f(\x)$.  In the special
case that $u=1$,  $C_{{\rho}, 1}( y f(\x))$ can be regarded as a
smooth alternative to the SVM hinge loss $[1-y f(\x)]_{+}$.
We refer to $C_{{\rho}, u}(y f(\x))$ as $C$-losses.

It is worth noting that $V_{1, 0}(z)$ is the logistic function and
$V_{{\rho}, 0}(z)$ has been
proposed by \citet{ZhangIR:2001} for binary logistic regression.
We keep in mind that $u\geq 0$ for $V_{{\rho}, u}(z)$ through this paper.

\subsection{Properties}

It is obvious that $C_{{\rho}, u}(z)$ and $V_{{\rho}, u}$ are infinitely smooth with respect to $z$.
Moreover, the first-order and second-order derivatives of  $C_{{\rho}, u}(z)$ with respect to $z$
are given as
\begin{align*}
C_{{\rho}, u}'(z) & = - \frac{u}{{\rho} \log[ 1{+} \exp(u/{\rho})]} \frac{\exp \frac{u {-}
z} {{\rho}}} {1{+} \exp \frac{u {-}
z} {{\rho}}}, \\
C_{{\rho}, u}''(z) & = \frac{u}{{\rho}^2 \log[ 1{+} \exp(u/{\rho})]} \frac{\exp \frac{u {-}
z} {{\rho}}} {(1{+} \exp \frac{u {-}
z} {{\rho}})^2}.
\end{align*}
Since $C_{{\rho}, u}''(z)>0$ for any $z\in \RB$, $C_{{\rho}, u}(z)$  as well as $V_{{\rho}, u}(z)$
are strictly convex in $z$, for fixed ${\rho}>0$ and $u> 0$.

We now investigate relationships among the coherence functions
and hinge losses. First, we have the following
properties.

\begin{proposition} \label{prop:2}
Let $V_{{\rho}, u}(z)$ and $C_{{\rho}, u}(z)$ be defined by (\ref{eq:b_c}) and (\ref{eq:b_closs}). Then,
\begin{enumerate}
\item[\emph{(i)}] \; $u {\times}I_{[z\leq 0]} \leq  [u{-} z]_{+}  \leq  V_{{\rho}, u}(z) \leq   {\rho} \log 2{+}[u {-} z]_{+}$;
\item[\emph{(ii)}] \; $\frac{1}{2}(u{-} z)  \leq  V_{{\rho}, u}(z) - {\rho} \log 2$;
\item[\emph{(iii)}] \; $\lim_{{\rho} \rightarrow 0}V_{{\rho},u}(z) = [u {-} z]_{+}$ \; and \; $\lim_{{\rho} \rightarrow \infty}V_{{\rho}, u}(z) - {\rho} \log 2 = \frac{1}{2}(u{-}
z)$;
\item[\emph{(iv)}] \; $u {\times}I_{[z\leq 0]} \leq C_{{\rho}, u}(z) \leq  V_{\rho, u}(z) $;
\item[\emph{(v)}] \;  $\lim_{{\rho} \rightarrow 0}  C_{{\rho}, u}(z) =  [u {-} z]_{+}$ and
$\lim_{{\rho} \rightarrow \infty}  C_{{\rho}, u}(z) = u$, for $u>0$.
\end{enumerate}
\end{proposition}

As a special case of $u=1$, we have $C_{{\rho}, 1}(z) \geq I_{[z\leq 0]}$. Moreover, $C_{{\rho}, 1}(z)$ approaches $(1{-}z)_{+}$ as
${\rho} \rightarrow 0$. Thus, $C_{{\rho}, 1}(z)$ is a majorization of $I_{[z\leq 0]}$.

As we mentioned earlier, $V_{{\rho}, 0}(z)$ are used
to devise logistic regression models. We can see from Proposition~\ref{prop:2} that
$V_{{\rho}, 0}(z) \geq [{-}z]_{+}$, which implies that a logistic regression model is possibly no longer a large-margin classifier.
Interestingly, however,  we consider a variant of $V_{{\rho}, u}(z)$ as
\[
L_{{\rho}, u}(z) = \frac{1}{\log(1+ \exp(u/{\rho}))}\log\big[ 1{+} \exp( (u- z)/{\rho}) \big], \; {\rho}>0, \; u\geq 0,
\]
which always satisfies that $L_{{\rho}, u}(z)\geq I_{[z\leq 0]}$ and $L_{{\rho}, u}(0)=1$,  for any $u\geq 0$. Thus, the $L_{{\rho}, u}(z)$ for ${\rho}>0$ and $u\geq 0$
are majorizations of $I_{[z\leq 0]}$. In particular, $L_{{\rho}, 1}(z) =C_{{\rho}, 1}(u)$ and $L_{1, 0}(z)$
is the logit function.

In order to explore the relationship of $C_{{\rho}, u}(z)$ with $(u{-}z)_{+}$, we now consider some properties of $L_{{\rho}, u}(z)$
when regarding it
respectively as a function of ${\rho}$ and of $u$.

\begin{proposition} \label{prop:variant} Assume ${\rho}>0$ and  $u\geq 0$. Then,
\begin{enumerate}
\item[\emph{(i)}] \; $L_{{\rho}, u}(z)$ is a deceasing function in ${\rho}$ if $z< 0$,
and it is an increasing function in ${\rho}$ if $z\geq 0$;
\item[\emph{(ii)}] \; $L_{{\rho},u}(z)$ is a deceasing function in $u$ if $z< 0$,
and it is an increasing function in $u$ if $z\geq 0$.
\end{enumerate}
\end{proposition}

\begin{figure} [!ht]
\begin{center}
  \begin{tabular}{cc} \hspace{-0.5cm}
   \includegraphics*[width=75mm, height=56mm]{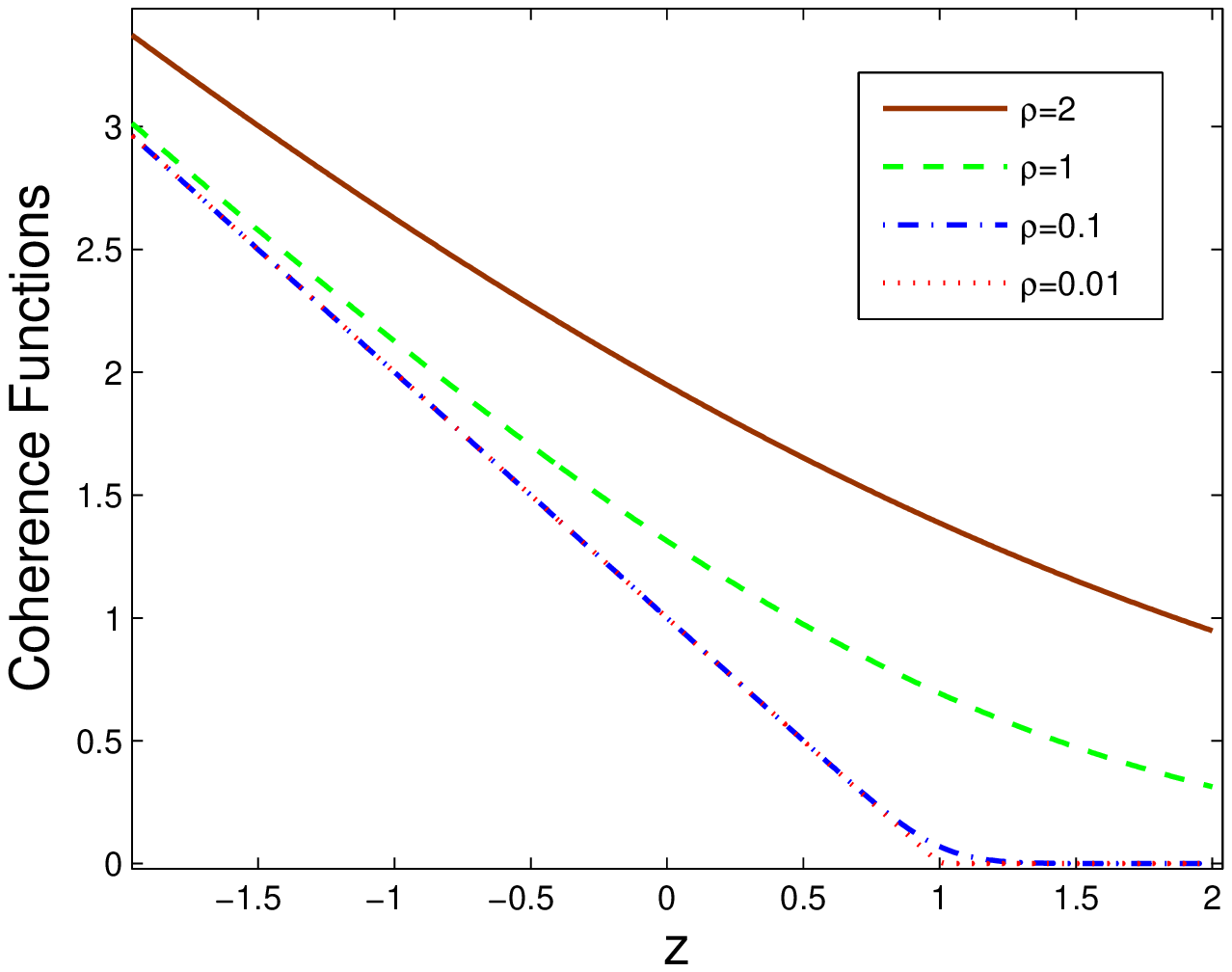} &\hspace{-0.1cm}
   \includegraphics*[width=75mm, height=56mm]{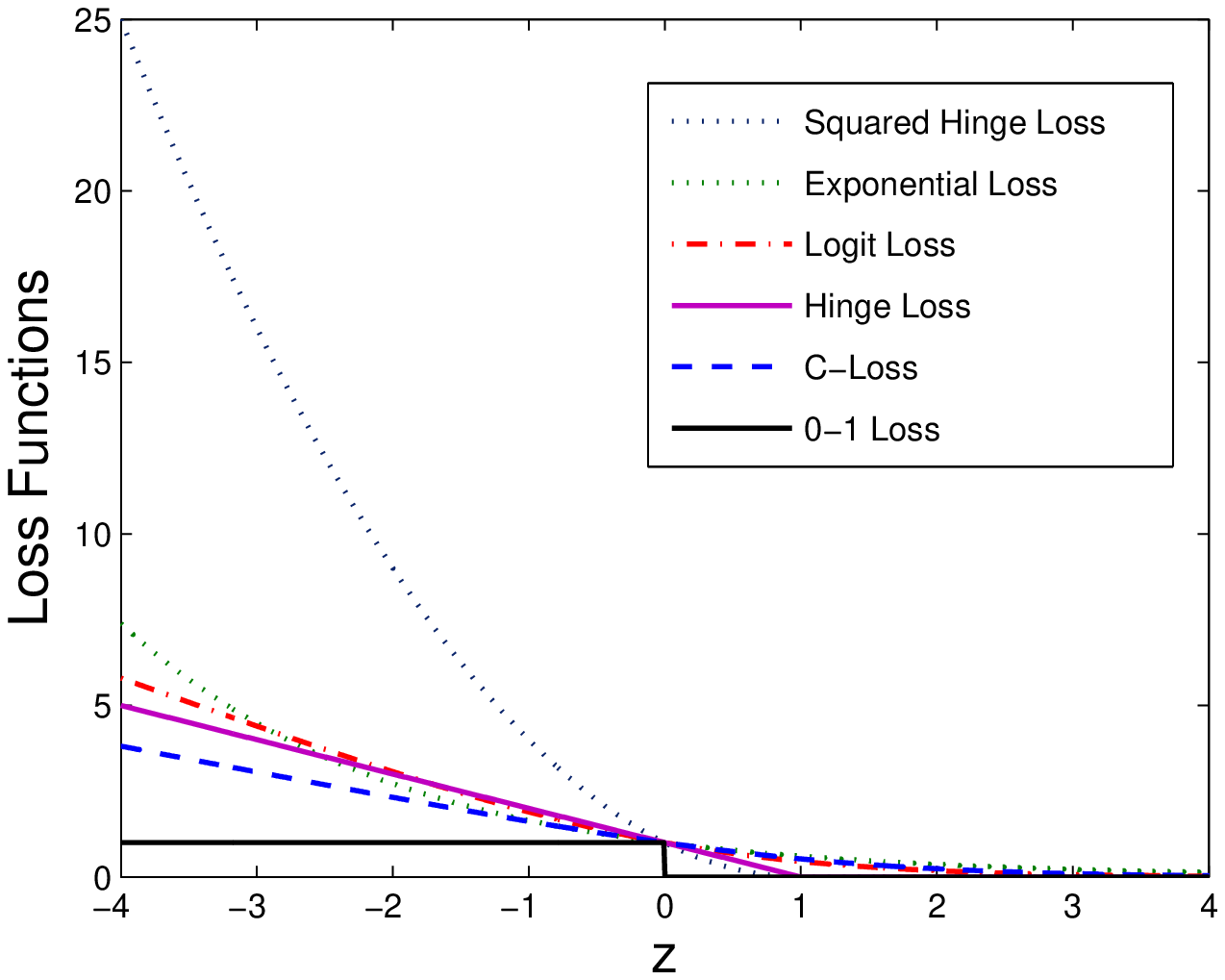} \hspace{-0.3cm} \\
  %  \hspace{-0.9cm}
    (a) & \hspace{-0.6cm}(b) \\
   \includegraphics*[width=75mm, height=56mm]{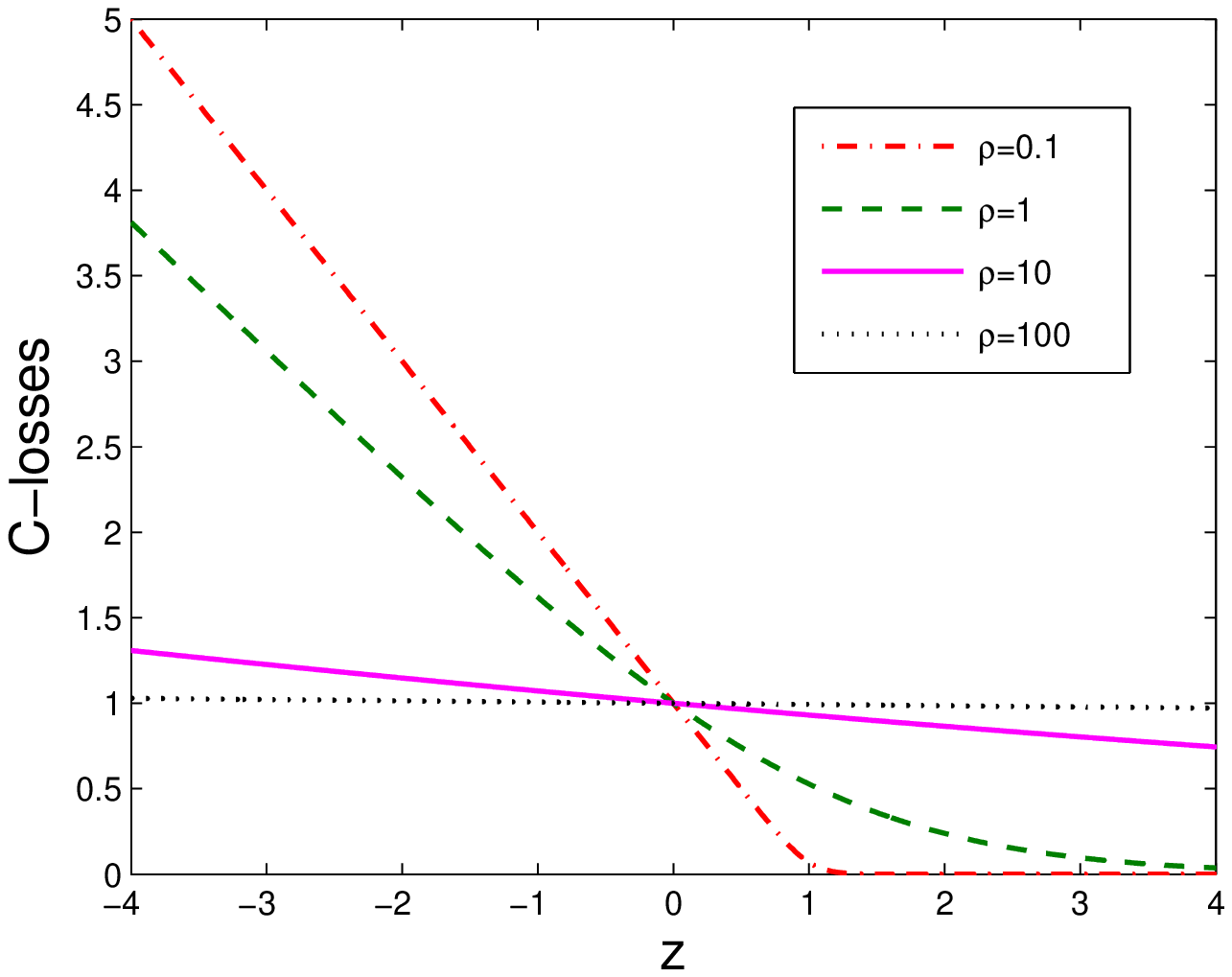} &\hspace{-0.1cm}
   \includegraphics*[width=75mm, height=56mm]{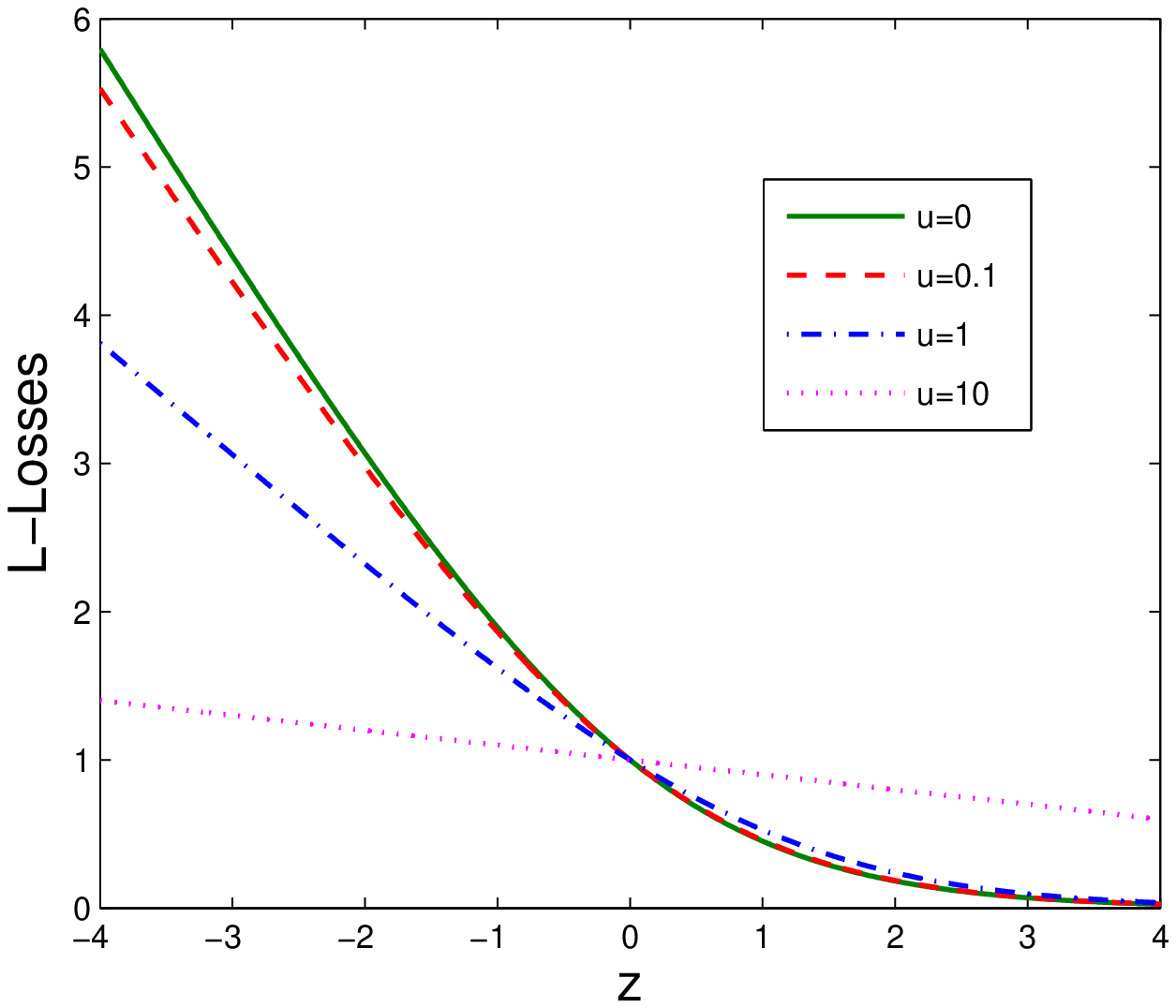} \hspace{-0.3cm} \\
  %  \hspace{-0.9cm}
    (c) & \hspace{-0.6cm}(d)
    \end{tabular}
\caption{These functions are regarded as a function of $z=y f(\x)$.
(a) Coherence functions $V_{{\rho}, 1}(z)$ with ${\rho}=0.01$, ${\rho}=0.1$,
${\rho}=1$ and ${\rho}=2$. (b) \; A variety of majorization loss functions,
 \emph{$C$-loss}: $C_{1, 1}(z)$; \emph{Logit loss}: $L_{1, 0}(z)$;
\emph{Exponential loss}: $\exp({-}z/2)$; \emph{Hinge loss}: $[1 {-} z]_{+}$; \emph{Squared Hinge Loss}: $([1 {-} z]_{+})^2$.
(c) \; $C_{{\rho}, 1}(z)$ (or $L_{{\rho}, 1}(z)$) with  ${\rho}=0.1$, ${\rho}=1$, ${\rho}=10$ and ${\rho}=100$ (see Proposition~\ref{prop:variant}-(i)).
(d) \; $L_{1, u}(z)$ with $u=0$, $u=0.1$, $u=1$ and $u=10$ (see Proposition~\ref{prop:variant}-(ii)).
} \label{fig:loss}
\end{center}
%\vspace{-0.1in}
\end{figure}

Results similar to those in Proposition~\ref{prop:variant}-(i) also apply to $C_{{\rho}, u}(z)$
because of $C_{{\rho}, u}(z)= u L_{{\rho}, u} (z)$. Then, according to Proposition~\ref{prop:2}-(v), we have
that $u = \lim_{{\rho} \rightarrow +\infty} C_{{\rho}, u}(z) \leq C_{{\rho}, u}(z)\leq  \lim_{{\rho} \rightarrow 0} C_{{\rho}, u}(z) = (u{-}z)_{+}$ if $z<0$ and $(u{-}z)_{+} = \lim_{{\rho} \rightarrow 0} C_{{\rho}, u}(z) \leq C_{{\rho}, u}(z)\leq \lim_{{\rho} \rightarrow +\infty} C_{{\rho}, u}(z) =u$ if $z\geq 0$.
It follows from Proposition~\ref{prop:variant}-(ii) that $C_{{\rho}, 1}(z) = L_{{\rho}, 1}(z) \leq L_{{\rho}, 0}(z)$ if $z<0$
and $C_{{\rho}, 1}(z) = L_{{\rho}, 1}(z) \geq L_{{\rho}, 0}(z)$ if $z\geq 0$.
In addition, it is easily seen that $(1-z)_{+} \geq ((1-z)_{+})^2$ if  $z\geq 0$ and $(1-z)_{+} \leq ((1-z)_{+})^2$
otherwise. We now obtain the following proposition:

\begin{proposition} \label{prop:u1} Assume ${\rho}>0$. Then,
$C_{{\rho}, 1}(z) \leq \min \big\{L_{{\rho}, 0}(z), \; [1{-}z]_{+}, \; ([1{-}z]_{+})^2 \big\}$ if $z<0$, and
$C_{{\rho}, 1}(z) \geq \max \big\{L_{{\rho}, 0}(z), \; [1{-}z]_{+}, \; ([1{-}z]_{+})^2  \big\}$
if $z\geq 0$.
\end{proposition}
This proposition is depicted in Figure~\ref{fig:loss}. 
Owing to the relationships of the $C$-loss
$C_{{\rho},1}(y f(x))$ with the hinge and logit losses, it is
potentially useful in devising new large-margin classifiers.

We now turn to the derivatives of $C_{{\rho}, u}(z)$ and $(u{-}z)_{+}$. It is immediately verified that $-1 \leq C'_{{\rho}, u}(z) \leq 0$.
Moreover, we have
\[
\lim_{{\rho} \rightarrow 0} C'_{{\rho}, u} (z) = \lim_{{\rho} \rightarrow 0} V'_{{\rho}, u} (z) = \left \{
\begin{array}{ll} 0 & z >u,  \\ - \frac{1}{2} & z= u, \\
-1 & z < u.
\end{array} \right.
\]
Note that $(u{-}z)_{+}'=-1$ if $z< u$ and $(u{-}z)_{+}'= 0$ if $z>u$. Furthermore, $\partial (u{-}z)_{+}|_{z=u}=[-1, 0]$
where $\partial (u{-}z)_{+} |_{z=u}$ denotes the
subdifferential of $(u{-}z)_{+}$ at $z=u$.  Hence,
\begin{proposition} \label{prop:dev1} For a fixed $u> 0$, we have
that $\lim_{{\rho} \rightarrow 0} C'_{{\rho}, u} (z) \in  \; \partial (u{-}z)_{+}$.
\end{proposition}

This proposition again establishes a connection of the hinge loss with the limit of $C_{{\rho}, u} (z)$ at ${\rho}=0$. Furthermore,
we obtain from Propositions~\ref{prop:2} and~\ref{prop:dev1} that  $\partial (u{-}z)_{+} = \partial \lim_{{\rho} \rightarrow 0} C_{{\rho}, u} (z) \ni  \lim_{{\rho} \rightarrow 0} \partial C_{{\rho}, u} (z)$.

\subsection{Consistency in Classification Methods}

We now apply the coherence function to the development of classification methods.
Recall that $C_{{\rho}, u}'(0)$ exists and is negative.
Thus, the $C$-loss $C_{{\rho}, u}(y f(x))$ is Fisher-consistent (or classification calibrated)~\citep{BartlettJordanMcauliffe:2006}.
In particular, we have the following theorem.

\begin{theorem} \label{thm:cohen}
Assume $0<\eta <1$ and $\eta \neq \frac{1}{2}$. Consider the optimization
problem
\[
\min_{f\in \RB} \; R(f, \eta) := V_{{\rho}, u}(f) \eta + V_{{\rho}, u}(-f) (1-\eta)
\]
for fixed ${\rho}>0$  and $u\geq 0$. Then, the minimizer is unique and is given by
\begin{equation} \label{eq:f_estimate}
 f_{*}(\eta) =   {\rho} \log \frac{(2\eta{-}1)\exp(\frac{u}{{\rho}}) + \sqrt{(1{-}2\eta)^2 \exp(\frac{2u}{{\rho}}) + 4 \eta(1{-}\eta)} }{2 (1-\eta)}.
\end{equation}
Moreover, we have $f_{*}>0$ if and only if $\eta> 1/2$.  Additionally, the inverse function $f_{*}^{-1}(f)$ exists and it is given by
\begin{equation} \label{eq:p1}
\tilde{\eta}(f) := f_{*}^{-1}(f) = \frac{1 + \exp (\frac{f -u} {{\rho}} )} {1 + \exp (-\frac{u+f} {{\rho}})
+ 1 + \exp (\frac{f -u} {{\rho}})}, \;\; \mbox{for } \; f \in \RB.
\end{equation}
\end{theorem}

The minimizer $f_{*}(\x)$
of $R(f(\x)):=\EB(V_{{\rho}, u}(Yf(X))|X=\x)$ and its inverse $\tilde{\eta}(\x)$ are immediately obtained by replacing $f$  with
$f(\x)$ in (\ref{eq:f_estimate}) and (\ref{eq:p1}). Since  for $u>0$ the minimizers
of $\EB(C_{{\rho}, u}(Yf(X))|X=\x)$ and $\EB({V}_{{\rho}, u}(Yf(X))|X =\x)$ are the same,
this theorem shows that ${C}_{{\rho}}(y f(\x), u)$ is also
Fisher-consistent.
We see from Theorem~\ref{thm:cohen} that the explicit expressions
of  $f_{*}(\x)$ and its inverse $\tilde{\eta}(\x)$ exist.
In the special case that  $u=0$, we have $f_{*}(\x) =  {\rho} \log \frac{\eta(\x)}{1-\eta(\x)}$ and $\tilde{\eta}(\x) =
\frac{1}{ 1+ \exp (-{f}(\x)/{\rho})}$. Furthermore,
when ${\rho}=1$, as expected, we recover logistic regression. In other words, the result is identical with that in Proposition~\ref{prop:1} for logistic regression.

We further consider properties of $f_{*}(\eta)$.  In particular, we have
the following proposition.

\begin{proposition} \label{thm:limit}
Let $f_{*}(\eta)$ be defined by (\ref{eq:f_estimate}). Then,
\begin{enumerate}
\item[\emph{(i)}] \; $\sign(f_{*}(\eta)) = \sign({\eta}{-}1/2)$. \item[\emph{(ii)}] \;
$\lim_{{\rho} \rightarrow 0}  f_{*}(\eta) = u {\times} {\sign}(\eta-1/2)$. %\quad for $u>0$.
\item[\emph{(iii)}] \; $f_{*}'(\eta) =\frac{d f_{*}(\eta)}{d \eta} \geq \frac{{\rho}}{\eta(1-\eta)}$ with equality if and only if $u=0$.
\end{enumerate}
\end{proposition}

Proposition~\ref{thm:limit}-(i) shows that the classification rule with $f_{*}(\x)$
is equivalent to the Bayes rule.  In the special case that $u=1$, we have from Proposition~\ref{thm:limit}-(ii) that $\lim_{{\rho} \rightarrow 0}  f_{*}(\x) = {\sign}(\eta(\x)-1/2)$. This implies that the
current $f_{*}(\x)$ approaches the solution of $\EB((1-Yf(X))_{+}|X=\x)$, which corresponds to the conventional SVM method (see Proposition~\ref{prop:1}).

We now treat $\tilde{\eta}(f)$ as a function of ${\rho}$. The following proposition is easily proven.

\begin{proposition} \label{prop:p_est} Let
$\tilde{\eta}(f)$ be defined by (\ref{eq:p1}). Then, for fixed $f \in \RB$ and $u> 0$,
 $\lim_{{\rho} \rightarrow \infty}  \tilde{\eta}(f) = \frac{1}{2}$ and
\[
\lim_{{\rho} \rightarrow 0}  \tilde{\eta}(f) = \left\{ \begin{array} {l@{\quad}l} 1 & \mbox{if } f >u, \\
\frac{2}{3} & \mbox{if } f =u, \\ \frac{1}{2} & \mbox{if } -u<f <u, \\
\frac{1}{3} & \mbox{if } f =-u, \\ 0 & \mbox{if } f <-u. \end{array} \right.
\]
\end{proposition}

As we discuss in the previous subsection, $V_{{\rho}, u}(z)$ is obtained when
setting $w=\exp((u{-}z)/{\rho})/(1+\exp((u{-}z)/{\rho}))$ by using the maximum entropy principle.
Let $z= y f(\x)$. We further write $w$ as $w_1(f) = 1/[1+\exp((f{-}u)/{\rho})]$ when $y=1$ and as
$w_2(f) = 1/[1+\exp(-(f{+}u)/{\rho})]$ when $y=-1$.

We now explore the relationship of $\tilde{\eta}(f)$ with $w_1(f)$ and $w_2(f)$. Interestingly, we first find that
\[
\tilde{\eta}(f) = \frac{w_2(f)}{w_1(f) + w_2(f)}.
\]
It is easily proven that $w_1(f) + w_2(f) \geq 1$ with equality if and only if $u=0$.
We thus have that $\tilde{\eta}(f) \leq  w_2(f)$, with equality if and only if $u=0$; that is, the loss becomes logit function $V_{{\rho}, 0}(z)$.
Note that $w_2(f)$ represents the probability of the event
$\{u+f > 0\}$ and $\tilde{\eta}(f)$ represents the probability
of the event $\{f> 0\}$. Since the event $\{f> 0\}$ is a subset
of the event $\{u+f> 0\}$, we have $\tilde{\eta}(f) \leq  w_2(f)$.
Furthermore, the statement that $\tilde{\eta}(f) =  w_2(f)$ if and only if $u=0$
is equivalent to $\{u+f> 0\}=\{f> 0\}$ if and only if $u=0$. This implies that only the logit loss induces
$\tilde{\eta}(f) =  w_2(f)$.

As discussed in Section~\ref{sec:surrogate}, $\tilde{\eta}(\x)$ can be regarded as
a reasonable estimate of the true class probability $\eta(\x)$. Recall that
$\triangle R(\eta, f) = R(\eta, f) - R(\eta, f_{*}(\eta))$ and
$\triangle R_{f} = \EB_{X} [\triangle R(\eta(X), f(X))]$ such that
$\triangle R_{f}$ can be viewed as the expected distance between
$\tilde{\eta}(\x)$ and $\eta(\x)$.

For an arbitrary fixed  $f \in \RB$, we have
\[
\triangle R(\eta, f) = R(\eta, f) - R(\eta, f_{*}(\eta)) = \eta {\rho} \log \frac{1+ \exp\frac{u{-}f}{{\rho}}}{1+ \exp\frac{u{-}f_{*}(\eta)}{{\rho}}}
+ (1-\eta) {\rho} \log \frac{1+ \exp\frac{u{+}f}{{\rho}}}{1+ \exp\frac{u{+}f_{*}(\eta)}{{\rho}}}.
\]
The first-order derivative of $\triangle R(\eta, f)$ with respect to $\eta$ is
\[
\frac{d \triangle R(\eta, f)} {d \eta}={\rho} \log \frac{1+ \exp\frac{u{-}f}{{\rho}}}{1+ \exp\frac{u{-}f_{*}(\eta)}{{\rho}}}
- {\rho} \log \frac{1+ \exp\frac{u{+}f}{{\rho}}}{1+ \exp\frac{u{+}f_{*}(\eta)}{{\rho}}}.
\]
The Karush-Kuhn-Tucker (KKT) condition for the minimization problem
is as follows:
\[
 \eta \frac{\exp \frac{u {-}
f_{*}(\eta) } {{\rho}}}{1{+} \exp \frac{u {-}
f_{*}(\eta)} {{\rho}}} + (1-\eta) \frac{\exp \frac{u {+}
f_{*}(\eta)} {{\rho}}}{ 1{+} \exp \frac{u {+}
f_{*}(\eta)} {{\rho}} } =0,
\]
and the second-order  derivative of $\triangle R(\eta, f)$ with respect to $\eta$ is given by
\[
\frac{d^2 \triangle R(\eta, f)} {d \eta^2}= \Big( \frac{1}{1+ \exp({-}\frac{u{-}f_{*}(\eta)}{{\rho}})}
+ \frac{1}{1+ \exp({-}\frac{u{+}f_{*}(\eta)}{{\rho}}) } \Big) f_{*}'(\eta) = \big[w_1(f_{*}(\eta))+ w_2(f_{*}(\eta)) \big] f_{*}'(\eta).
\]
According to Proposition~\ref{thm:limit}-(iii) and using $w_1(f_{*}(\eta))+ w_2(f_{*}(\eta))\geq 1$, we have
\[
\frac{d^2 \triangle R(\eta, f)} {d \eta^2} \geq \frac{{\rho}}{\eta(1-\eta)},
\]
with equality if and only if $u=0$. This implies $\frac{d^2 \triangle R(\eta, f)} {d \eta^2} >0$. Thus, for a fixed $f$,
$\triangle R(\eta, f)$ is strictly convex in $\eta$. Subsequently, we have that $\triangle R(\eta, f)\geq 0$
with equality $\eta=\tilde{\eta}$, or equivalently, $f=f_{*}$.

Using the Taylor expansion of $\triangle R(\eta, f)$ at $\tilde{\eta}:=\tilde{\eta}(f) = f_{*}^{-1}(f)$,
we thus obtain a lower bound for $\triangle R(\eta, f)$; namely,
\begin{align*}
\triangle R(\eta, f) & =\triangle R(\tilde{\eta}, f) - \frac{d \triangle R(\tilde{\eta}, f)} {d \eta} (\eta-\tilde{\eta})
 +\frac{1}{2} \frac{d^2 \triangle R(\bar{\eta}, f)} {d \eta^2} (\eta-\tilde{\eta})^2 \\
 & = \frac{1}{2} \frac{d^2 \triangle R(\bar{\eta}, f)} {d \eta^2} (\eta-\tilde{\eta})^2
 \geq \frac{{\rho}}{2\bar{\eta}(1-\bar{\eta})} (\eta-\tilde{\eta})^2 \geq 2 {\rho} (\eta-\tilde{\eta})^2,
\end{align*}
where $\bar{\eta} \in (\tilde{\eta}, \eta) \subset [0, 1]$. In particular, we have that $\triangle R(\eta, 0) \geq 2{\rho} (\eta-0.5)^2$.
According to Theorem~2.1 and Corollary~3.1 in \citet{ZhangAS:2004}, the following theorem is immediately established.

\begin{theorem} \label{thm:est} Let $\epsilon_1= \inf_{f(\cdot) \in \FM} \EB_{X} [\triangle R(\eta(X), f(X))]$,  and let $f_{*}(\x) \in \FM$ such that
\[\EB_{X}[R(\eta(X), f_{*}(X))]  \leq \inf_{f(\cdot) \in \FM} \EB_{X}[R(\eta(X), {f}(X))] + \epsilon_2
\]
for $\epsilon_2 \geq 0$. Then for $\epsilon=\epsilon_1+\epsilon_2$,
\[
\triangle R_{f_{*}} = \EB_{X} [\triangle R(\eta(X), f_{*}(X))] \leq \epsilon
\]
and
\[
\Psi_{f_{*}} \leq \hat{\Psi} + \big(2\epsilon /{\rho} \big)^{1/2},
\]
where  $\Psi_{f_{*}}= \EB_{P} I_{[Y f_{*}(X)\leq 0]}$, and $\hat{\Psi}=\EB_{P} I_{[Y(2\eta(X){-}1)\leq 0]}$ is the optimal Bayes error.
\end{theorem}

\subsection{Analysis}

For notational simplicity, we will use $C_{{\rho}}(z)$ for
$C_{{\rho}, 1}(z)$.
%Suppose that we are given  $\rho>0$.
Considering  $f(\x) = \alpha+ \bet'\k$,
we define  an regularized optimization problem of
the form
\begin{equation} \label{eq:creg}
\min_{\alpha, \; \bet} \Big\{ \frac{1}{n} \sum_{i=1}^{n} C_{{\rho}}(y_if(\x_i)) + \frac{\gamma_n}{2} \bet' \K \bet \Big \}.
\end{equation}
%for $j=1, \ldots, m$.
Here we assume that the regularization parameter $\gamma$ relies on
the number $n$ of training data points, thus we denote it by $\gamma_n$.

Since the optimization problem (\ref{eq:creg}) is convex with respect
to $\alpha$ and $\bet$,
the solution exists and is unique. Moreover, since $C_{\rho}$ is infinitely smooth,  we can resort to the Newton-Raphson
method to solve (\ref{eq:creg}).

\begin{proposition} \label{pro:c_svm} Assume that  $\gamma_n$  in (\ref{eq:creg}) and $\gamma$ in (\ref{eq:h_svm0}) are same.  Then
the minimizer of (\ref{eq:creg}) approaches the minimizer of (\ref{eq:h_svm0}) as ${\rho} \rightarrow 0$.
\end{proposition}

This proposition is obtained directly from Proposition~\ref{prop:dev1}. For a fixed ${\rho}$,
we are also concerned with the universal consistency of the classifier based on (\ref{eq:creg}) with and without the offset term $\alpha$.

\begin{theorem} \label{thm:consistent2}
Let $K(\cdot, \cdot)$ be a universal kernel on $\XM{\times}\XM$.
Suppose we are given such a positive sequence  $\{\gamma_n\}$ that $\gamma_n \rightarrow 0$.
If
\[
n \gamma^2_n / \log n \rightarrow \infty,
\]
then the classifier based on (\ref{eq:creg}) is strongly universally consistent.
If
\[
n \gamma^2_n \rightarrow \infty,
\]
then the classifier based on (\ref{eq:creg}) with $\alpha=0$ is universally consistent.
\end{theorem}

\subsection{Class Probability Estimation of  SVM Outputs}
\label{sec:cpe}

As discussed earlier, the limit of the coherence function, $V_{{\rho},1}(y f(\x))$,
at ${\rho}=0$ is just the hinge loss. Moreover,  Proposition~\ref{thm:limit} shows that the minimizer of $V_{{\rho},1}(f)\eta+ V_{{\rho},1}(-f)(1{-}\eta)$
approaches that of $H(f)\eta+ H(-f)(1{-}\eta)$ as ${\rho}\rightarrow 0$. Thus,
Theorem~\ref{thm:cohen}
provides us with an approach to the estimation of the class probability for the conventional SVM.

In particular,
let $\hat{f}(\x)$ be the solution of the optimization problem (\ref{eq:h_svm0}) for the conventional SVM. In terms of Theorem~\ref{thm:cohen},
we suggest that the estimated class probability $\hat{\eta}(\x)$
is defined as
\begin{equation} \label{eq:p_svm0}
%\[
\hat{\eta}(\x) = \frac{1 + \exp (\frac{\hat{f}(\x) -1} {{\rho}} )} {1 + \exp (-\frac{1+\hat{f}(\x)} {{\rho}})
+ 1 + \exp (\frac{\hat{f}(\x) -1} {{\rho}} )}.
%\]
\end{equation}
Proposition~\ref{thm:limit} would seem to motivate setting $\rho$ to
a very small value in (\ref{eq:p_svm0}).  However, as shown in
Proposition~\ref{prop:p_est}, the probabilistic outputs degenerate to
$0$, $1/3$, $1/2$, $2/3$ and $1$ in this case.  Additionally, the classification
function $\hat{f}(\x)=\hat{\alpha}+ \sum_{i=1}^n \hat{\beta}_i K(\x, \x_i)$ is obtained via fitting a conventional SVM
model on the training data.  Thus, rather than attempting to specify
a fixed value of $\rho$ via a theoretical argument, we instead view
it as a hyperparameter to be fit empirically.

In particular, we fit ${\rho}$ by minimizing the generalized
Kullback-Leibler divergence (or cross-entropy error) between
$\hat{\eta}(X)$ and $\eta(X)$, which is given by
\[
\mbox{GKL}(\eta, \hat{\eta}) = \EB_{X} \Big[\eta(X) \log\frac{\eta(X)}{\hat{\eta}(X)} + (1{-}\eta(X)) \log \frac{1{-}\eta(X)}{1{-}\hat{\eta}(X)}  \Big].
\]
%Since  $\EB[(Y{+}1)/2|X] =\eta (X)$,
Alternatively,
we formulate the optimization problem for obtaining ${\rho}$ as
\begin{equation} \label{eq:eklo}
\min_{{\rho} > 0} \; \mbox{EKL}(\hat{\eta}) :=
- \frac{1}{n} \sum_{i=1}^n \Big\{\frac{1}{2}(y_i+1) \log \hat{\eta}(\x_i) + \frac{1}{2}(1-y_i) \log(1- \hat{\eta}(\x_i)) \Big\}.
\end{equation}
The problem can be solved by the Newton method.  In summary, one first
obtains $\hat{f}(\x)=\hat{\alpha}+ \sum_{i=1}^n \hat{\beta}_i K(\x, \x_i)$ via the conventional SVM model, and estiamtes
$\rho$ via the optimization problem in~(\ref{eq:eklo}) based on the
training data; one then uses the formula in (\ref{eq:p_svm0}) to
estimate the class probabilities for the training samples as well
as the test samples.

%%%%%%%%%%%%%%%%%%%%%%%%%%%%%%%%%%%%%%%%%%%%%%%%%%%%%%%%%%%%%%%%%%%%%%%%%%%%%%%%%%
%%%%%%%%%%%%%%%%%%%%%%%%%%%%%%%%%%%%%%%%%%%%%%%%%%%%%%%%%%%%%%%%%%%%%%%%%%%%%%%%%%
\section{${\cal C}$-Learning} \label{sec:c_learn}

%%%%%%%%%%%%%%%%%%%%%%%%%%%%%%%%%%%%%%%%%%%%%%%%%%%%%%%%%%%%%%%%%%%%%%%%%%%%%%%%%%
%%%%%%%%%%%%%%%%%%%%%%%%%%%%%%%%%%%%%%%%%%%%%%%%%%%%%%%%%%%%%%%%%%%%%%%%%%%%%%%%%%

Focusing on  the relationships of the $C$-loss $C_{{\rho}}(y f(x))$ (i.e., $C_{{\rho},1}(y f(x))$) with the hinge and logit losses,
we illustrate its application  in the construction
of large-margin classifiers.
Since $C_{\rho}(y f(\x))$ is smooth,
it does not tend to yield a sparse classifier.
However, we can employ a sparsification penalty $J(h)$
to arrive at sparseness.  We use the elastic-net
penalty of~\cite{ZouENET:2005} for the experiments
in this section.  Additionally, we study two forms
of $f(\x)$: kernel expansion and feature expansion.
Built on these two expansions, sparseness can subserve  the selection of support vectors and the selection of
features, respectively.  The resulting classifiers are called \emph{${\cal C}$-learning}.

\subsection{The Kernel Expansion}

In the kernel expansion approach, given a reproducing kernel
$K(\cdot, \cdot):$ $\XM \times \XM \rightarrow \RB$,  we define
the kernel expansion as   $f(\x) = \alpha + \sum_{i=1}^n \beta_i K(\x_i, \x)$ and
solve the following optimization problem:
\begin{equation} \label{eq:ker_c2}
\min_{\alpha, \bet} \frac{1}{n}
\sum_{i=1}^n
C_{{\rho}}(y_i f(\x_i)) + \gamma \Big((1{-}\omega) \frac{1}{2} \bet' \K \bet  + \omega \|\bet\|_{1} \Big),
\end{equation}
where  $\K=[K(\x_i, \x_j)]$ is the $n{\times}n$ kernel matrix.

It is worth pointing out that the current penalty is slightly different from the conventional elastic-net
penalty, which is  $(1{-}\omega) \frac{1}{2} \bet' \bet  + \omega \|\bet\|_{1}$.
In fact, the optimization problem~(\ref{eq:ker_c2}) can be viewed equivalently as the optimization problem
\begin{equation} \label{eq:ker_c3}
\min_{\alpha, \bet} \frac{1}{n}
\sum_{i=1}^n
C_{{\rho}}(y_i f(\x_i)) +  \frac{\gamma}{2} \bet' \K \bet
\end{equation}
under  the $\ell_1$ penalty $ \|\bet\|_{1}$. Thus, the method derived from (\ref{eq:ker_c2})
enjoys the generalization  ability of the
conventional kernel supervised learning method derived from (\ref{eq:ker_c3})  but also the sparsity of the $\ell_1$ penalty.

Recently, \citet{FriedmanHastieTibshirani08:2008} devised a pathwise coordinate descent algorithm for regularized
logistic regression problems
in which the elastic-net penalty is  used.
In order to solve the optimization problem in (\ref{eq:ker_c2}), we employ this pathwise coordinate descent algorithm.

Let the current estimates of $\alpha$ and $\beta$ be $\hat{\alpha}$ and $\hat{\beta}$. We first form a quadratic approximation to $\frac{1}{n}
\sum_{i=1}^n
C_{{\rho}}(y_i f(\x_i))$, which is
\begin{equation} \label{eq:quadratic_approximation_kernel_binary}
{Q}(\alpha, \bet) = \frac{1}{2n \rho} \sum_{i=1}^n q(\x_i)(1-q(\x_i))\big(\alpha
+ \k_i' \bet - z_i \big)^2 + \mbox{Const},
\end{equation}
where
\begin{eqnarray*}
z_i  &=&  \hat{\alpha} + \k_i' \hat{\bet} + \frac{{\rho}}{y_i(1- q(\x_i))}, \\
q(\x_i) &=& \frac{\exp[(1-y_i (\hat{\alpha} + \k_i' \hat{\bet})/{\rho}]}{1 + \exp[(1-y_i (\hat{\alpha} + \k_i' \hat{\bet}))/{\rho}]},\\
\k_i &=& (K(\x_1,\x_i),\ldots,K(\x_n,\x_i))'.
\end{eqnarray*}

We then employ  coordinate descent
to solve the weighted least-squares problem as follows:
\begin{equation} \label{obj:Binary_Kernel_least-squares}
\min_{\alpha, \bet} \; G(\alpha, \bet):=  {Q} (\alpha, \bet)+\gamma \Big((1{-}\omega) \frac{1}{2} \bet' \K \bet  + \omega \|\bet\|_{1} \Big).
\end{equation}
Assume that we have estimated $\tilde{\bet}$ for $\bet$ using $G(\alpha, \bet)$. We now set
$\frac{\partial G(\alpha,\tilde{\bet})}{\partial \alpha} = 0$ to find
the new estimate of $\alpha$:
\begin{equation}\label{eq:binary_a}
\tilde{\alpha} = \frac{\sum_{i=1}^n q(\x_i)(1-q(\x_i))\big(z_i - \k_i' \tilde{\bet} \big)}{\sum_{i=1}^n q(\x_i)(1-q(\x_i))}.
\end{equation}

On the other hand, assume that we have estimated $\tilde{\alpha}$ for $\alpha$ and $\tilde{\beta}_l$ for $\beta_l$ $(l=1,\ldots,n, l\neq j)$.
We now optimize ${\beta}_j$. In particular, we only consider the gradient at $\beta_j\neq 0$. If $\beta_j> 0$, we have
\[
\frac{\partial G(\tilde{\alpha}, \tilde{\bet})}{\partial \beta_j}=\frac{1}{n{\rho}}
\sum_{i=1}^n K_{ij}q(\x_i)(1-q(\x_i))\big(\alpha {+} \k_i' \tilde{\bet} {-} z_i \big) + \gamma(1{-}\omega) (K_{jj} \beta_j  {+} \k_j \check{\bet}) + \gamma\omega
\]
and, hence,
\begin{equation} \label{eq:binary_b}
\tilde{\beta_j} =
\frac{S(t-\gamma(1-\omega)\k_j'\check{\bet},
\gamma\omega)}{\frac{1}{n{\rho}} \sum_{i=1}^n K_{ij}^2q(\x_i)(1-q(\x_i))+\gamma(1-\omega)K_{jj}},
\end{equation}
where $t=\frac{1}{n \rho} \sum_{i=1}^n K_{ij}q(\x_i)(1-q(\x_i))\big(z_i - \tilde{\alpha}
- \k_i' \check{\bet}\big)$, $\check{\bet} = (\tilde{\beta}_1,\ldots,\tilde{\beta}_{j-1},0,\tilde{\beta}_{j+1},\ldots,\tilde{\beta}_{n})'$, $K_{ij} = K(\x_i,\x_j)$,
and
$S(\mu,\nu)$ is the soft-thresholding operator:
\begin{align*}
S(\mu,\nu) & = \sign(\mu)(|\mu|-\nu)_+ \nonumber\\
& = \left  \{ \begin{array}{ll} \mu-\nu & \mbox{if } \mu>0 \mbox{ and } \mu<|\nu| \\
\mu+\nu & \mbox{if } \mu<0 \mbox{ and } \mu<|\nu|
\\
0 & \mbox{if } \mu>|\nu|. \end{array} \right.\label{eq:hardprob}
\end{align*}
Algorithm~\ref{alg:Binary_C_learning}
summarizes the coordinate descent algorithm for binary $C$-learning.

\begin{algorithm}[tb]
   \caption{The coordinate descent algorithm for binary ${\cal C}$-learning}
   \label{alg:Binary_C_learning}
\begin{algorithmic}
   \STATE {\bfseries Input:} $\TM = \{\x_i,y_i\}_{i=1}^n$, $\gamma$, $\omega$, $\epsilon_m$, $\epsilon_i$, ${\rho}$;
   \STATE {\bfseries Initialize: $\tilde{\alpha} = \alpha_0$, $\tilde{\bet} = \bet_0$}
   \REPEAT
   \STATE Calculate $G(\tilde{\alpha}, \tilde{\bet})$ using (\ref{obj:Binary_Kernel_least-squares});
   \STATE $\alpha^\star\leftarrow \tilde{\alpha}$;
   \STATE $\bet^\star\leftarrow \tilde{\bet}$;
   \REPEAT
   \STATE $\bar{\alpha} \leftarrow \tilde{\alpha}$;
   \STATE $\bar{\bet} \leftarrow \tilde{\bet}$;
   \STATE Calculate $\tilde{\alpha}$ using~(\ref{eq:binary_a});
   \FOR{$j=1$ {\bfseries to} $n$}
   \STATE Calculate $\tilde{\beta}_j$ using~(\ref{eq:binary_b});
   \ENDFOR
   \UNTIL $\|\tilde{\alpha}-\bar{\alpha}\|+\|\tilde{\bet}-\bar{\bet}\|<\epsilon_i$
   \UNTIL $\|\tilde{\alpha}-\alpha^\star\|+\|\tilde{\bet}-\bet^\star\|<\epsilon_m$
   \STATE {\bfseries Output:} $\tilde{\alpha}$, $\tilde{\bet}$, and $f(\x)= \tilde{\alpha}+ \sum_{i=1}^n K(\x_i, \x) \tilde{\beta_i}$.
\end{algorithmic}
\end{algorithm}

\subsection{The Linear Feature Expansion}

In the linear feature expansion approach, we let $f(\x) = a + \x' \b$, and pose the
following optimization problem:
\begin{equation} \label{eq:lin_conh2}
\min_{a, \b}  \frac{1}{n}
\sum_{i=1}^n
C_{{\rho}}(y_i f(\x_i)) + \gamma J_{\omega}(\b),
\end{equation}
where for $\omega \in[0, 1]$
\begin{align*}
J_{\omega}(\b) & = (1-\omega) \frac{1}{2} \|\b\|^2_2 + \omega
\|\b\|_1  = \sum_{j=1}^d \Big[ \frac{1}{2}(1-\omega) b_j^2 + \omega
|b_j| \Big].
\end{align*}
The elastic-net penalty maintains the sparsity of the $\ell_1$ penalty, but the number of variables to be selected
is no longer bounded by $n$. Moreover, this penalty tends to generate similar coefficients for highly-correlated
variables.
We  also use a coordinate descent algorithm to solve the
optimization problem~(\ref{eq:lin_conh2}). The algorithm is similar to
that for the kernel  expansion and the details are omitted here.

%%%%%%%%%%%%%%%%%%%%%%%%%%%%%%%%%%%%%%%%%%%%%%%%%%%%%%%%%%%%%%%%%%%%%
%%%%%%%%%%%%%%%%%%%%%%%%%%%%%%%%%%%%%%%%%%%%%%%%%%%%%%%%%%%%%%%%%%%%%
\section{Experimental Results} \label{sec:examp}

In Section~\ref{sec:examp0} we  report the results of experimental evaluations 
of our method for  class probability estimation  of the conventional SVM given  
in Section~\ref{sec:cpe}.  In Section~\ref{sec:expe2} we present results for 
the $\CM$-learning method given in Section~\ref{sec:c_learn}.

%%%%%%%%%%%%%%%%%%%%%%%%%%%%
%%%%%%%%%%%%%%%%%%%%%%%%%%%%%%%%%%%%%%%%%
%%%%%%%%%%%%%%%%%%%%%%%%%%%%%%%%%%%%%%%%%%%%%%%%%%%%%%%%%%%%%%%%%%%%%
\subsection{Simulation for Class Probability Estimation of SVM Outputs} \label{sec:examp0}

We validate our estimation method for the class probability of SVM outputs 
(``Ours for SVM"), comparing it with several alternatives:  Platt's 
method~\citep{Platt:1999}, Sollich's method~\citep{SollichML:2002}, and the method  
of~\cite{WangShen:2008} (WSL's). 
Since penalized (or regularized) logistic regression (PLR) and 
$\CM$-learning can directly calculate  class probability,
we also implement them. Especially, the class probability of $\CM$-learning 
outputs is  based on (\ref{eq:p1}) where we set $\rho=1$ and $u=1$
since $\CM$-learning itself employs the same setting.   

%and Zhu \& Hastie's method~\citep{ZhuPLR:2004}.

We  conducted our analysis over two simulation datasets which were used by \cite{WangShen:2008}.
The first simulation dataset, $\{(x_{i1}, x_{i2}; y_i)\}_{i=1}^{1000}$, was 
generated as follows. The $\{(x_{i1}, x_{i2})\}_{i=1}^{1000}$
were uniformly sampled from a unit disk $\{(x_{1}, x_{2}): x^2_{1}+ x^2_{2}\leq 1\}$.
Next, we set $y_i=1$ if $x_{i1}\geq 0$ and $y_{i}=-1$ otherwise, $i=1, \ldots, 1000$.
Finally, we randomly chose $20\%$ of the samples and flipped their labels.  
Thus, the true class probability $\eta(Y_i=1|x_{i1}, x_{i2})$ was either 0.8 or 0.2.

The second dataset, $\{(x_{i1}, x_{i2}; y_i)\}_{ i=1} ^{1000}$, was generated 
as follows. First, we randomly assigned $1$ or $-1$ to $y_i$ for  $i=1, \ldots, 1000$
with equal probability. Next, we generated $x_{i1}$ from the uniform distribution 
over $[0, 2\pi]$, and set $x_{i2}= y_i(\sin(x_{i1}) +\epsilon_i)$ where
$\epsilon_i \sim N(\epsilon_i|1, 0.01)$.  For the data, the true class 
probability of $Y=1$ was given by
\[
\eta(Y=1|x_{1}, x_{2})  =  \frac{N(x_{2}|\sin(x_{1}) {+}1, 0.01)}
{N(x_{2}|\sin(x_{1}){+}1, 0.01)+N(x_{2}|-\sin(x_{1}) {-}1, 0.01)}.
\]

The simulation followed the same setting as that in \cite{WangShen:2008}. 
That is, we randomly selected 100 samples for training and the remaining 
900 samples for test.  We did 100 replications for each dataset. 
The values of generalized Kullback-Leibler loss (GKL) and classification 
error rate (CER) on the test sets were averaged over these 100 simulation replications. 
Additionally, we employed a Gaussian RBF kernel  $K(\x_i, \x_j) = \exp(-\|\x_i-\x_j\|^2/\sigma^2)$
where the parameter $\sigma$ was set as the median  distance between the positive and negative classes.
%Finally,  our method sets the initial value of ${\rho}$ as ${\rho}_0 = 10^2$.
We reported GKL and CER as well as the corresponding standard deviations in Tables~\ref{tab:gkl1} and \ref{tab:cer1}
in which the results with  the PLR method, the tuned Platt method and the WSL method 
are directly cited from \cite{WangShen:2008}.

Note that  %for PLR  the value of GKL is infinity when the estimated probability becomes exactly 0 or 1,  so  
the results with PLR were averaged only over 66 nondegenerate replications~\citep{WangShen:2008}. Based on GKL and CER, 
the performance of $\CM$-learning is the best in these two simulations.  
With regard to GKL, our method for SVM outperforms the original and tuned versions of Platt's method as well as the method of \cite{WangShen:2008}. 
Since our estimation method is based on the $\hat{\eta}(\x)$ in (\ref{eq:p_svm0}), the CER with this class probability $\hat{\eta}(\x)$ is identical to that with the
conventional  SVM.  This also applies to  Sollich's method, thus we did not include the CER with this method. However, Table~\ref{tab:cer1} shows  that this does not necessarily hold  for   Platt's method for SVM probability outputs.
%\citep{Platt:1999,SollichML:2002,WangShen:2008}. 
In other words,  $\hat{\eta}(\x)>1/2$ is not equivalent to $\hat{f}(\x)>0$ for Platt's method. In fact,
\cite{Platt:1999} used this sigmoid-like function to improve the classification accuracy of  the conventional SVM. As for the method of \cite{WangShen:2008} which
is built  on a sequence of weighted classifiers,  the CERs of the method should be different from those of the original SVM.
With regard to CER, the performance of PLR  is the worst in most cases.
%Obviously, the learnt parameter $\rho$ is very stable over each dataset.

\begin{figure} [!ht]
\begin{center}
  \begin{tabular}{cc}\hspace{-0.9cm}
 \includegraphics*[width=80mm, height=60mm]{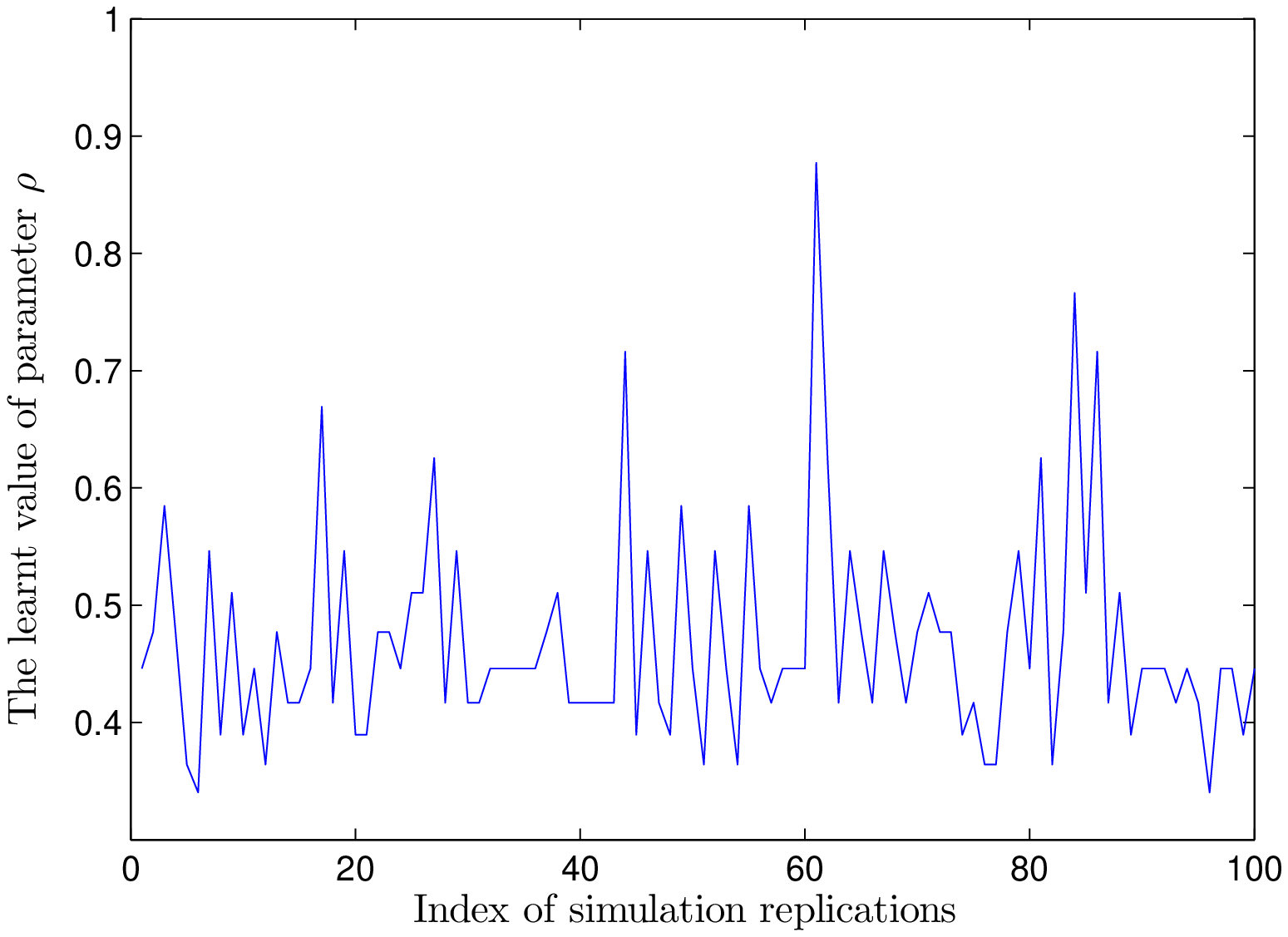} &\hspace{-0.9cm}
 \includegraphics*[width=80mm, height=60mm]{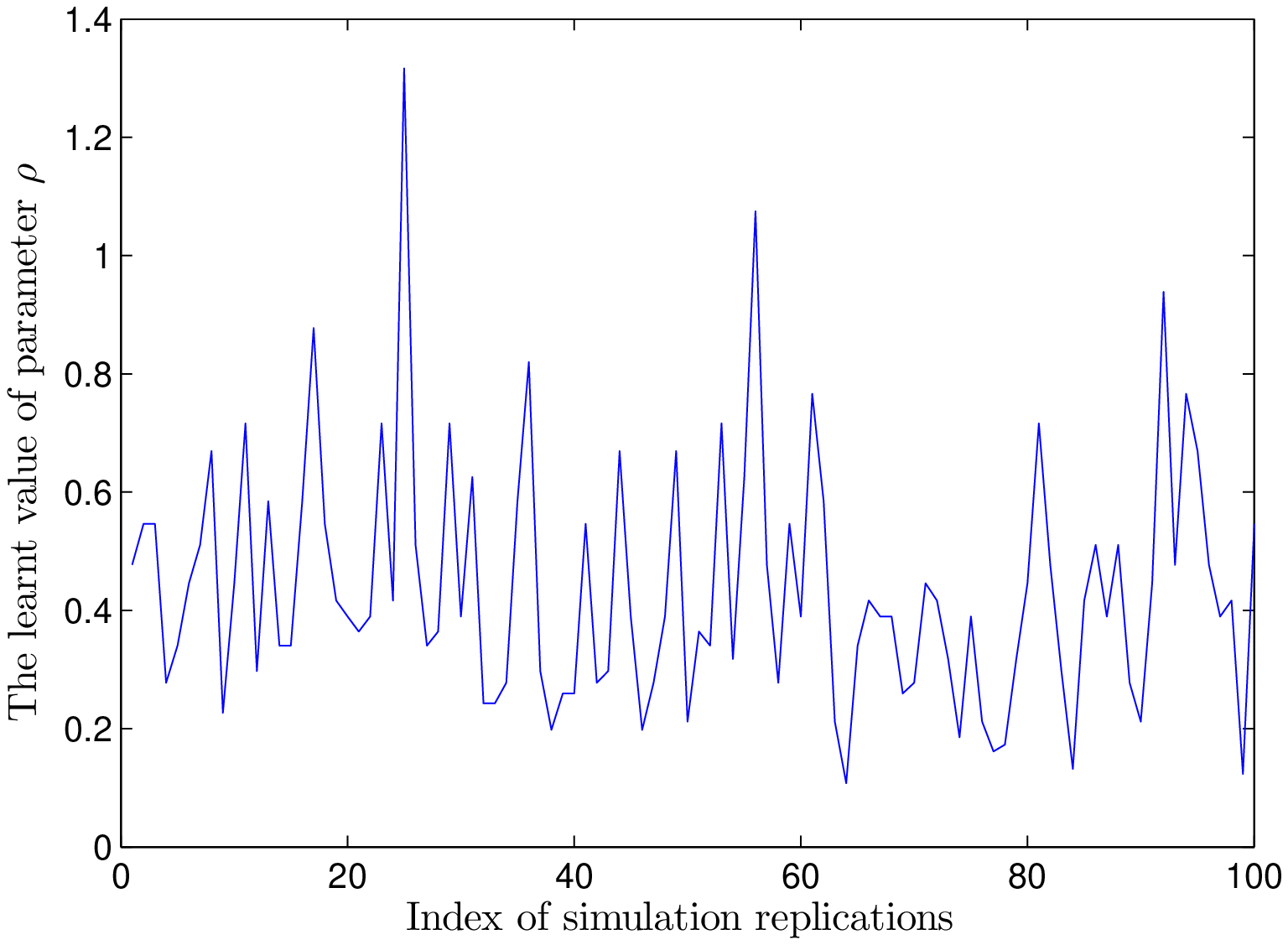}\\\hspace{-0.9cm}
  (a) Simulation  1 &\hspace{-0.9cm} (b) Simulation  2
   \end{tabular}
\caption{The learned values of  parameter $\rho$ vs.  simulation replications in our estimation method for the class probability of SVM outputs.}
\label{fig:rho_value}
\end{center}
%\vspace{-0.10in}
\end{figure}

In addition, Figure~\ref{fig:rho_value} plots the estimated values of   parameter $\rho$ with respect to the 100 simulation replications in our method for class probability estimation of the original SVM. 
For simulation 1, the estimated values of $\rho$ range  from 0.3402 to 0.8773, while
they range from 0.1077 to 1.3166 for simulation 2. 

%Sample 2:
%Our: TE 0.065(+_0.0015), 
%PlattÕs: GKL 0.163(+_0.0018), TE0.077(+_0.0024)
%
%Sample 1:
%PlattÕs: GKL 0. 582(+_0.0035), TE 0. 234(+_0.0026)
%Our: TE 0.219(+_0.0023),

%The C learning:
%simulation 1:
%GKL 0.549(+_0.0016), TE 0.214(+-0.0015)
%Simulation 2
%GKL 0.134(+_0.0014), TE 0.061(+_0.0019)
%Sollich's method data 1 0.566(+_0.0021)
%data 2 0.155(+_0.0017)

\begin{table}[!ht]  \setlength{\tabcolsep}{1.3pt}
\caption{Values of GKL over the two simulation test sets
(standard deviations are shown in parentheses).}
\label{tab:gkl1} %\vspace{0.05cm}
\begin{center}
\begin{tabular}{c||c|c|c|c |c| c|c}
\hline
 & {PLR} & Platt's &{Tuned Platt} &
{WSL's} & Sollich's &{Ours for SVM} & {$\CM$-learning}\\
\hline
\raisebox{-1.1ex}{Data 1}  &  0.579 & 0.582  & 0.569   & 0.566                          & 0.566  &  0.558  & 0.549 \\
           &  ($\pm 0.0021$) &  ($\pm 0.0035$) &   ($\pm 0.0015$) &  ($\pm 0.0014$) & ($\pm 0.0021$) & ($\pm 0.0015$)  & ($\pm 0.0016$)  \\
           \hline
\raisebox{-1.1ex}{Data 2}   & 0.138 & 0.163   & 0.153 & 0.153 & 0.155  &  0.142  & 0.134 \\
             & ($\pm 0.0024$)  & ($\pm 0.0018$)& ($\pm 0.0013$)  & ($\pm 0.0010$)  & ($\pm 0.0017$) &  ($\pm 0.0016$) & ($\pm 0.0014$) \\ \hline
\end{tabular}
\end{center}
%\vskip -0.1in
\end{table}
\begin{table}[!ht]
\caption{Values of CER  over the two simulation test sets
(standard deviations are shown in parentheses).}
\label{tab:cer1} %\vspace{0.05cm}
\begin{center}
\begin{tabular}{c||c|c | c  | c |c}
\hline
  & PLR &{Platt's} &
{WSL's} &{Ours for SVM} & {$\CM$-learning} \\
\hline
\raisebox{-1.1ex}{Data 1}  & 0.258                   & 0.234                    & 0.217 &  0.219 & 0.214  \\
             &  ($\pm 0.0053$)  &   ($\pm 0.0026$) &  ($\pm 0.0021$) &  ($\pm 0.0021$) & ($\pm 0.0015$) \\ 
\hline
\raisebox{-1.1ex}{Data 2}   & 0.075   & 0.077 & 0.069 &  0.065  & 0.061 \\ 
            & ($\pm 0.0018$) &  ($\pm 0.0024$)  &  ($\pm 0.0014$)  &  ($\pm 0.0015$) & ($\pm 0.0019$)  \\   \hline
\end{tabular}
\end{center}
%\vskip -0.1in
\end{table}

\subsection{The Performance Analysis of $\CM$-Learning}
\label{sec:expe2}

To evaluate the performance of our $\CM$-learning method, we further conducted empirical studies on
several benchmark datasets and compared $\CM$-learning with two closely
related classification methods: the hybrid huberized SVM (HHSVM) of~\citet{LWangICML:2007} and
the regularized logistic regression model (RLRM) of~\citet{FriedmanHastieTibshirani08:2008}, both with the
elastic-net penalty. All the three classification methods
were implemented in both the feature and  kernel expansion settings.

In the experiments we used 11 binary classification datasets.
Table~\ref{tab:data} gives a summary of these  benchmark datasets.
The seven binary datasets of digits were obtained from the publicly available  USPS dataset of handwritten
digits as follows. The first six datasets were generated from the digit pairs
\{(1, 7), (2, 3), (2, 7), (3, 8), (4, 7), (6,9)\}, and 200 digits were chosen within each class of each
dataset is 200. The USPS (odd vs.\ even) dataset consisted of the first 80 images per digit in the USPS training set.

The two  binary artificial datasets of ``g241c" and ``g241d"
were generated via the setup presented by~\citet{OChapellekSSL:2006}.
Each class of these two datasets consisted of 750 samples.

The two binary gene datasets of ``colon'' and ``leukemia'' were also used in our experiments. The
``colon'' dataset, consisting of 40 colon tumor samples and 22
normal colon tissue samples with 2,000 dimensions, was obtained by employing an Affymetrix
oligonucleotide array to analyze more than 6,500
human genes expressed in sequence tags~\citep{AlonPNAS:1999}.
The ``leukemia'' dataset is of the same type as the ``colon'' cancer dataset~\citep{GolubScience:1999}, and it was obtained with respect to two variants of leukemia, i.e., acute myeloid leukemia (AML) and acute lymphoblastic leukemia (ALL). It initially contained expression levels of 7129 genes taken over 72 samples (AML, 25 samples, or ALL, 47 samples), and then it was pre-feature selected, leading to a feature space with 3571 dimensions.

In our experiments, each dataset was randomly partitioned into two disjoint subsets
as the training and test, with the percentage of the training data samples also 
given in Table~\ref{tab:data}.  Twenty random partitions were chosen for each dataset,
and the average and standard deviation of their classification error rates
over the test data were reported.

\begin{table}[!ht]\setlength{\tabcolsep}{10.3pt}
%\vspace{-0.15in}
\caption{Summary of the benchmark datasets: $m$---the number of classes; $d$---the dimension
of the input vector;  $k$---the size of the
dataset; $n$---the number of the
training data.}
\label{tab:data}
\begin{center}
%\begin{footnotesize}
\begin{tabular}{l|cccc}
\hline
Dataset & $m$ & $d$  & k &  $n/k$ \\
\hline
USPS (1 vs. 7)    & 2     & 256   & 400   & 3\% \\
USPS (2 vs. 3)    & 2     & 256   & 400   & 3\% \\
USPS (2 vs. 7)    & 2     & 256   & 400   & 3\% \\
USPS (3 vs. 8)    & 2     & 256   & 400   & 3\% \\
USPS (4 vs. 7)    & 2     & 256   & 400   & 3\% \\
USPS (6 vs. 9)    & 2     & 256   & 400   & 3\% \\
USPS (Odd vs. Even)    & 2     & 256   & 800  & 3\%\\
g241c             & 2     & 241   & 1500  & 10\%\\
g241d             & 2     & 241   & 1500  & 10\%\\
colon & 2 & 2000 & 62 & 25.8\%\\
leukemia  & 2 & 3571 & 72 & 27.8\%\\
%duke-breast-cancer & 2 & 7129 & 44& 22.7\%\\
\hline
\end{tabular}
%\end{footnotesize}
\end{center}
\end{table}

\begin{table}[!ht]\setlength{\tabcolsep}{10.3pt}
\vspace{-0.1in} \caption{Classification error rates (\%) and
standard deviations on the 11 datasets for the feature expansion setting.}
\label{tab:linear}
\begin{center}
%\begin{footnotesize}
\begin{tabular}{l|cccr}
\hline
Dataset & HHSVM & RLRM & $\CM$-learning \\
\hline \hline
(1 vs. 7)         & 2.29$\pm$1.17 & 2.06$\pm$1.21 & \textbf{1.60$\pm$0.93} \\
(2 vs. 3)         & \textbf{8.13$\pm$2.02} & 8.29$\pm$2.76 & 8.32$\pm$2.73 \\
(2 vs. 7)         & 5.82$\pm$2.59 & 6.04$\pm$2.60 & \textbf{5.64$\pm$2.44} \\
(3 vs. 8)         &12.46$\pm$2.90 &\textbf{10.77$\pm$2.72} &11.74$\pm$2.83 \\
(4 vs. 7)         &7.35$\pm$2.89 & 6.91$\pm$2.72 & \textbf{6.68$\pm$3.53} \\
(6 vs. 9)         & 2.32$\pm$1.65 & 2.15$\pm$1.43 & \textbf{2.09$\pm$1.41} \\
(Odd vs. Even)    &20.94$\pm$2.02 &19.83$\pm$2.82 &\textbf{19.74$\pm$2.81} \\
\hline
g241c                  &22.30$\pm$1.30 &21.38$\pm$1.12 &\textbf{21.34$\pm$1.11} \\
g241d                  &24.32$\pm$1.53 &\textbf{23.81$\pm$1.65} &23.85$\pm$1.69 \\
colon                  &14.57$\pm$1.86 &14.47$\pm$2.02 &\textbf{12.34$\pm$1.48}\\
leukemia               &4.06$\pm$2.31 &4.43$\pm$1.65 &\textbf{3.21$\pm$1.08} \\
%duke-breast-cancer     &18.66$\pm$3.33 &17.90$\pm$3.24 &\textbf{17.71$\pm$3.03}\\
\hline \hline
\end{tabular}
%\end{footnotesize}
\end{center}
\vspace{-0.20in}
\end{table}

\begin{table}[bt]\setlength{\tabcolsep}{10.3pt}
\vspace{-0.15in} \caption{Classification error rates (\%) and
standard deviations on the 11 datasets for the RBF kernel setting.}
\label{tab:kernel}
\begin{center}
%\begin{footnotesize}
\begin{tabular}{l|cccr}
\hline
Dataset & HHSVM & RLRM & $\CM$-learning \\
\hline \hline
(1 vs. 7)         & 1.73$\pm$1.64 & 1.39$\pm$0.64 & \textbf{1.37$\pm$0.65} \\
(2 vs. 3)         & 8.55$\pm$3.36 & 8.45$\pm$3.38 & \textbf{8.00$\pm$3.32} \\
(2 vs. 7)         & 5.09$\pm$2.10 & 4.02$\pm$1.81 & \textbf{3.90$\pm$1.79} \\
(3 vs. 8)         &12.09$\pm$3.78 &10.58$\pm$3.50 &\textbf{10.36$\pm$3.52} \\
(4 vs. 7)         &6.74$\pm$3.39 & 6.92$\pm$3.37 & \textbf{6.55$\pm$3.28} \\
(6 vs. 9)         & 2.12$\pm$0.91 & 1.74$\pm$1.04 & \textbf{1.65$\pm$0.99} \\
(Odd vs. Even)    &28.38$\pm$10.51&26.92$\pm$6.52 &\textbf{26.29$\pm$6.45} \\
\hline
g241c                  &\textbf{21.38$\pm$1.45} &21.55$\pm$1.42 &21.62$\pm$1.35 \\
g241d                  &25.89$\pm$2.15 &22.34$\pm$1.27 &\textbf{20.37$\pm$1.20} \\
colon                  &14.26$\pm$2.66 &14.79$\pm$2.80 &\textbf{13.94$\pm$2.44}\\
leukemia               &2.77$\pm$0.97 &2.74$\pm$0.96 &\textbf{2.55$\pm$0.92}\\
%breast cancer     &\textbf{19.64$\pm$2.56} &20.76$\pm$2.02 &20.19$\pm$2.39\\
\hline\hline
\end{tabular}
%\end{footnotesize}
\end{center}
\vspace{-0.15in}
\end{table}

%Proposition~\ref{prop:variant} shows that as $T\rightarrow 0$, $\CM_T(\g(\x), c)$  approaches ${\max} \{g_j(\x){+}1 {-}
%\IB_{[j=c]}\} {-} g_c(\x)$.  However,
%when $T$ gets very small, it can lead to numerical problems and often
%makes the algorithm unstable.
Although we can seek an optimum $\rho$ using computationally intensive methods such  as cross-validation,
the experiments showed that when $\rho$ takes a value in $[0.1, 2]$, our method is always able to
obtain promising performance.
Here our reported results are based on the setting of $\rho=1$, due to the relationship
of the $\CM$-loss $\CM(z)$ with the hinge loss $(1-z)_{+}$
and the logit loss $\log(1+\exp(-z))$ (see our analysis in Section~\ref{sec:c_loss} and Figure~\ref{fig:loss}).

As for the parameters $\gamma$ and $\omega$, they were selected by cross-validation
for all the classification methods.
In the kernel expansion, the RBF kernel $K(\x_i,\x_j) = \exp(-\|\x_i-\x_j\|^2/\sigma^2)$ was employed,
and $\sigma$ was set to the
mean Euclidean distance among the input samples. For $\CM$-learning, the other parameters were set
as follows: $\epsilon_m=\epsilon_i=10^{-5}$.

Tables~\ref{tab:linear} and~\ref{tab:kernel} show the test results corresponding
to the linear feature expansion and RBF kernel expansion, respectively. From the tables, we can see that for
the overall performance of $\CM$-learning is slightly better than the two competing methods
in the feature and  kernel settings generally.

\begin{figure} [!ht]
\begin{center}
  \begin{tabular}{cc}\hspace{-0.9cm}
 \includegraphics*[width=80mm, height=60mm]{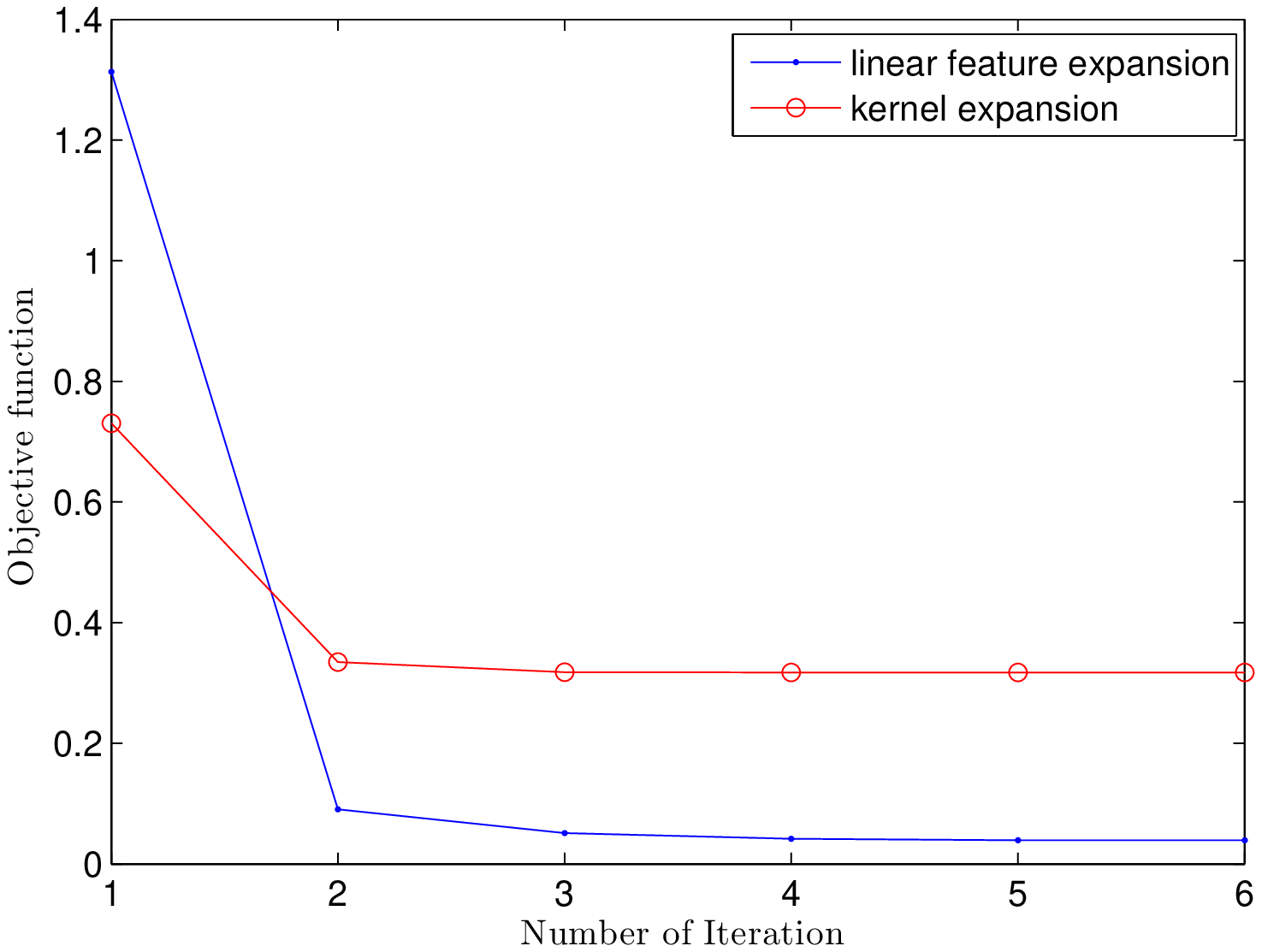} &\hspace{-0.9cm}
 \includegraphics*[width=80mm, height=60mm]{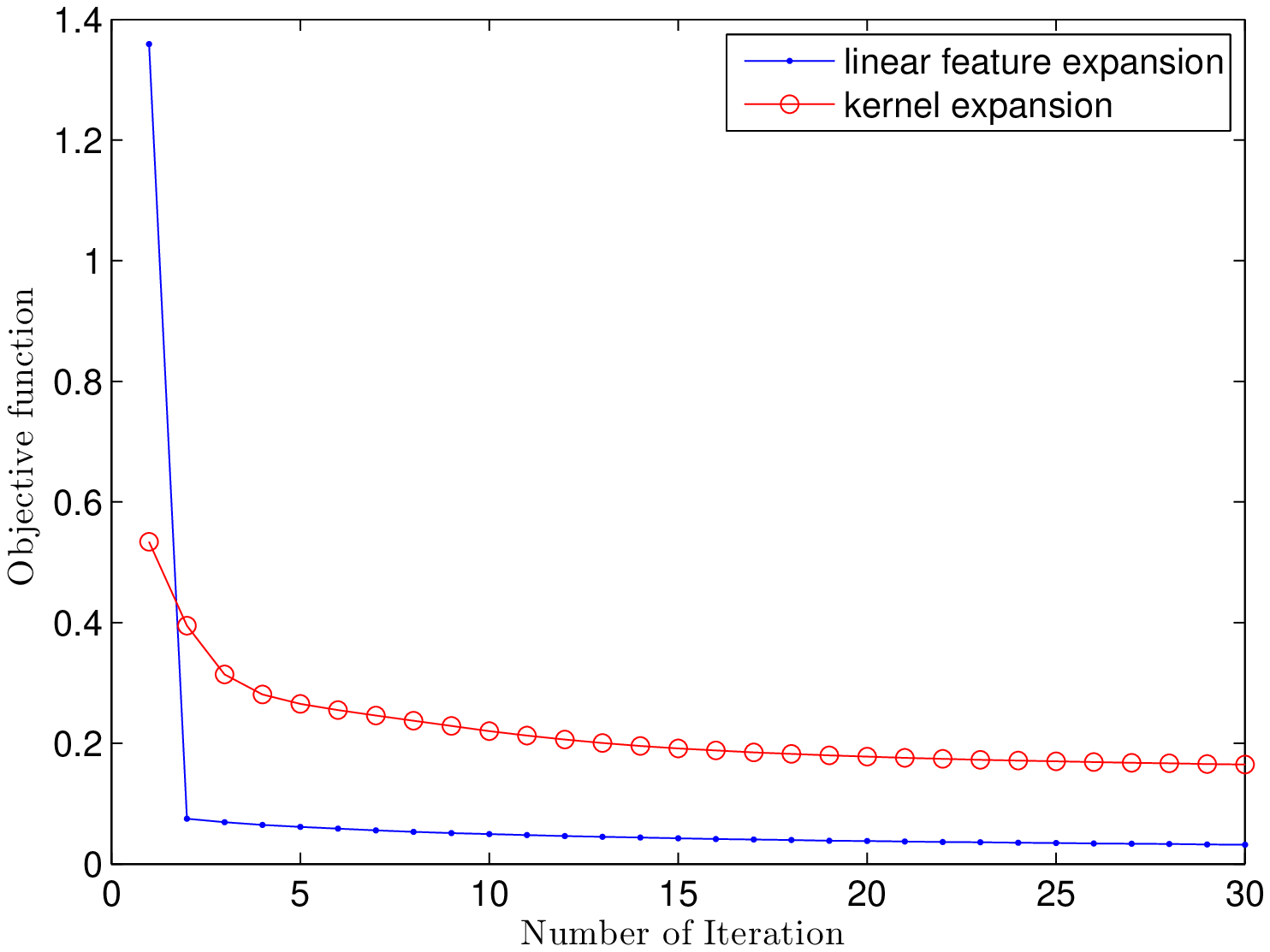}\\\hspace{-0.9cm}
  (a) &\hspace{-0.9cm} (b)
   \end{tabular}
\caption{Change of the value of the objective function for the $\CM$-learning as
the number of iterations in the coordinate descent procedure
increases in the linear feature  and RBF kernel cases on the dataset USPS (1 vs. 7):
(a) the  values of the objective function  (\ref{eq:ker_c2}) in the outer iteration;
(b) the objective function values  $G(\alpha, \bet)$
for the fature and RBF kernel cases in the inner iteration.}
\label{fig:obj_value}
\end{center}
%\vspace{-0.10in}
\end{figure}

Figure~\ref{fig:obj_value} reveals that the values of the objective functions for the linear feature and  RBF kernel versions
in the outer and inner
iterations tend to be significantly reduced as the number of iterations in the coordinate descent procedure
increases. Although we report only the change of the values of the objective function
for the dataset USPS (1 vs. 7) similar results were found on all other datasets.
This shows that the coordinate descent algorithm is very efficient.

We also conducted a systematic study of sparseness from the elastic-net penalty.
Indeed, the elastic-net penalty does give rise to sparse solutions for our $\CM$-learning methods.
Moreover, we found that similar to other methods the sparseness of the solution is
dependent on the parameters $\gamma$ and  $\omega$ that
were set to different values for different datasets using cross validation.

%\begin{footnotesize}
%\begin{figure}[htb]
%\begin{center}
%\begin{tabular}{c}\hspace{-0.8cm}
%\subfigure[]{\epsfig{figure=image/Fig_inner.eps,width=5cm}}\hspace{-0.55cm}
%\subfigure[]{\epsfig{figure=image/Fig_inner.eps,width=5cm}}%\hspace{-1cm}
%\end{tabular}
%\end{center}\vspace{-0.6cm}
%\caption{Change of leave-one-out error for SS-KFDA as the number of
%iterations in the gradient-descent procedure increases: (a)~case~I; (b)~case~II.} \label{Fig:leave_one_out_error}
%%\vspace{-0.3cm}
%\end{figure}
%\end{footnotesize}
%\vspace{-0.1cm}

%%%%%%%%%%%%%%%%%%%%%%%%%%%%%%%%%%%%%%%%%%%%%%%%%%%%%%%%%%%%%%%%%%%
%%%%%%%%%%%%%%%%%%%%%%%%%%%%%%%%%%%%%%%%%%%%%%%%%%%%%%%%%%%%%%%%%%%
\section{Conclusions} \label{sec:concl}

In this paper we have studied a family of  coherence functions and
considered the relationship between coherence functions and hinge functions.
In particular, we have established some important properties of these functions,
which lead us to a novel approach for class probability
estimation in the conventional SVM.  Moreover, we have proposed
large-margin classification methods using the $C$-loss function and the elastic-net penalty,
and developed  pathwise coordinate descent algorithms for parameter estimation.
We have theoretically established the Fisher-consistency of our classification methods and
empirically tested the classification performance on several benchmark datasets.
Our approach
establishes an interesting link between SVMs and logistic regression models due
to the relationship of the $C$-loss with the hinge and logit losses.

%Since the coherence function is smooth, it is naturally applied to multi-class 
%large-margin classification methods.  In fact, in a related line of
%work~\citep{ZhangAISTAT:2009} we have devised an efficient and effective
%multi-class boosting algorithm using a multi-class coherence function.
%
%
%\section*{Acknowledgments}
%
%The authors would like to thank the Action Editor and three anonymous referees 
%for their constructive comments and suggestions on the original version of this paper.
%This work has supported in part by the US ARL and the US ARO under contract/grant 
%number W911NF-11-1-0391.

%We will study the convergence rate of our  classification methods in future work.

%We introduce the elastic-net penalty into ${\cal C}$-learning to achieve sparseness,
%and devise
%coordinate descent algorithms for computationally efficient estimation under our ${\cal C}$-learning methods.

%%%%%%%%%%%%%%%%%%%%%%%%%%%%%%%%%%%%%%%%%%%%%%%%%%%%%%%%%%%%%%%%%%%%%%%%%%%%%%%%%%%%%%%%%%%%%%%%%%
%%%%%%%%%%%%%%%%%%%%%%%%%%%%%%%%%%%%%%%%%%%%%%%%%%%%%%%%%%%%%%%%%%%%%%%%%%%%%%%%%%%%%%%%%%%%%%%%%%
\appendix

%%%%%%%%%%%%%%%%%%%%%%%%%%%%%%%%%%%%%%%%%%%%%%%%%%%%%%%%%%%%%%%%%%%%%%%%%%%%%%%%%%%%%%%%%%%%%%%%%
%%%%%%%%%%%%%%%%%%%%%%%%%%%%%%%%%%%%%%%%%%%%%%%%%%%%%%%%%%%%%%%%%%%%%%%%%%%%%%%%%%%%%%%%%%%%%%%%%
\section{The Proof of Proposition~\ref{prop:2}}  \label{ap:a}

First, we have
\begin{align*}
{\rho} \log 2 + [u-z]_{+} - V_{{\rho}, u}(z)
 = {\rho} \log  \frac{ 2 \exp \frac{[u{-}z]_{+}} {{\rho}} } {1{+} \exp \frac{u {-}
z} {{\rho}}} \geq 0.
\end{align*}
Second, note that
\begin{align*}
%& {\rho} \log\Big[ 1{+} \sum_{j \neq c} \exp \frac{1 {+}
%g_j(\x) {-} g_c(\x) } {{\rho}} \Big] - {\rho}\log m - \frac{1}{m} \sum_{j\neq c} (1{+}g_j(\x)-g_c(\x))
{\rho} \log 2 + \frac{u-z}{2} - V_{{\rho}, u}(z) & = {\rho} \log
\frac{ 2 \exp \frac{1}{2}  \frac{u{-}z}
{{\rho}} } {1{+}  \exp \frac{u {-}
z} {{\rho}}} \\
& \leq {\rho} \log \frac{ \exp \frac{u {-}
z} {{\rho}} } {1{+}  \exp \frac{u {-}
z} {{\rho}}} \leq 0,
\end{align*}
where we use the fact that $\exp(\cdot)$ is convex.

%%%%%%%%%%%%%%%%%%%%%%%%%%%%%%%%%%%%%%%%%%%%%%%%%%%%%%%%%%%%%%%%%%%%%%%%%%%%%%%%%%%%%%%%%%%%%%%%%
%%%%%%%%%%%%%%%%%%%%%%%%%%%%%%%%%%%%%%%%%%%%%%%%%%%%%%%%%%%%%%%%%%%%%%%%%%%%%%%%%%%%%%%%%%%%%%%%%
%\section{Proof of Proposition~\ref{pro:wj}}  \label{ap:c}

Third, it immediately follows from Proposition~(i) that $\lim_{{\rho}
\rightarrow 0} V_{{\rho}, u}(z) = [u-z]_{+}$. Moreover, it is easily obtained that
\begin{align*}
\lim_{{\rho} \rightarrow \infty} V_{{\rho}, u}(z) -{\rho} \log 2 & =
\lim_{{\rho} \rightarrow \infty} \frac{ \log \frac { {1{+}  \exp \frac{u {-} z} {{\rho}}} } {2} }  {
\frac{1}{{\rho}}}
= \lim_{\alpha \rightarrow 0} \frac{ \log \frac { 1{+}  \exp \alpha (u {-}z)} {2} }  {\alpha} \\
& = \lim_{\alpha \rightarrow 0} \frac{ \frac{1}{2}
[u {-} z] \exp [\alpha (u {-} z)] } {\frac { 1{+} \exp \alpha (u {-} z)
} {2}}  = \frac{1}{2} (u {-} z).
\end{align*}

Since $\log(1+a)\geq \log(a)$ for $a>0$, we have
\[
 \frac{u}{\log[ 1{+} \exp(u/{\rho})] }\log\big[ 1{+} \exp \frac{u {-}
z} {{\rho}} \big] \leq \frac{u}{u/{\rho}}\log\big[ 1{+} \exp \frac{u {-}
z} {{\rho}} \big] = {\rho}  \log\big[ 1{+} \exp \frac{u {-}
z} {{\rho}} \big].
\]
We now consider that
\begin{align*}
\lim_{{\rho} \rightarrow \infty} C_{{\rho}, u}(z) = u \lim_{{\rho}
\rightarrow \infty}  \frac{\log\Big[ 1{+}  \exp
\frac{u {-} z} {{\rho}} \Big]}{\log[ 1{+}
\exp(u/{\rho})] } = u.
\end{align*}
Finally, since
\[
\lim_{\alpha \rightarrow \infty} \frac{\log[ 1{+}  \exp(u
\alpha)] }{\alpha u} =\lim_{\alpha \rightarrow \infty}\frac{
\exp(u \alpha)]}{1{+}  \exp(u \alpha)}=1 \; \mbox{ for } \;
u>0,
\]
we obtain $\lim_{{\rho} \rightarrow 0} C_{{\rho}, u}(z) =
[u-z]_{+}$. %However, we have $\lim_{{\rho} \rightarrow 0}
%C_{{\rho}, 0}(z) = + \infty$ whenever $z<0$ and $\lim_{{\rho} \rightarrow 0}
%C_{{\rho}, 0}(z) = 0$ if $z>0$.

%Third, the derivative of $V_{{\rho}, u}(\g(\x), c)$ w.r.t. ${\rho}$ is given
%by
%\begin{align*}
%\frac{\partial V}{\partial {\rho}} &= \log\Big[1+ \sum_{j\neq c}
%\exp\frac{u{+}g_j(\x){-}g_c(\x)}{{\rho}} \Big] - \frac{\sum_{j\neq c}
%\exp\frac{u{+}g_j(\x){-}g_c(\x)}{{\rho}} \frac{u{+}g_j(\x){-}g_c(\x)}{{\rho}}}
%{ 1+ \sum_{j\neq c} \exp\frac{u{+}g_j(\x){-}g_c(\x)}{{\rho}}} \\
%&= \max_{j}  \frac{u{+}g_j(\x){-}g_c(\x){-}u I_{[j=c]}}{{\rho}}
%-\frac{\sum_{j=1}^m \exp\frac{u{+}g_j(\x){-}g_c(\x){-}u
%I_{[j=c]}}{{\rho}} \frac{u{+}g_j(\x){-}g_c(\x){-}u I_{[j=c]}}{{\rho}}}
%{\sum_{j=1}^m \exp\frac{u{+}g_j(\x){-}g_c(\x){-}u I_{[j=c]}}{{\rho}}}
%\\
%& \geq 0.
%\end{align*}
%Thus, $V_{{\rho}, u}(\g(\x), c)$ is an increasing function w.r.t. ${\rho}$.

%%%%%%%%%%%%%%%%%%%%%%%%%%%%%%%%%%%%%%%%%%%%%%%%%%%%%%%%%%%%%%%%%%%%%%%%%%%%%%%%%%%%%%%%%%%%%%%%%
%%%%%%%%%%%%%%%%%%%%%%%%%%%%%%%%%%%%%%%%%%%%%%%%%%%%%%%%%%%%%%%%%%%%%%%%%%%%%%%%%%%%%%%%%%%%%%%%%
\section{The Proof of Proposition~\ref{prop:variant}}  \label{ap:d}

Before we prove Proposition~\ref{prop:variant}, we establish the
following lemma.

\begin{lemma} \label{lem:1} Assume that $x>0$, then $f_1(x)= \frac{x}{1+x} \frac{\log
x}{\log(1+x)}$ and $f_2(x)= \frac{x}{1+x} \frac{1}{\log(1+x)}$ are
increasing  and deceasing, respectively.
\end{lemma}
\begin{proof} The first derivatives of $f_1(x)$ and $f_2(x)$ are
\begin{align*}
f_1'(x) &= \frac{1}{(1+x)^2 \log^2(1+x) } \Big[\log x \log(1+x) +
\log(1+x) + x \log (1+x) - x \log x \Big] \\
f_2'(x) & =\frac{1}{(1+x)^2 \log^2(1+x) } [\log(1+x) -x] \leq 0.
\end{align*}
This implies that $f_2(x)$ is deceasing.  If $\log x\geq 0$, we have
$x \log (1+x)- x \log x\geq 0$. Otherwise, if $\log x <0$, we have $\log
x [\log (1+x) - x] \geq 0$. This implies that $f_1'(x)\geq 0$ is always
satisfied. Thus, $f_1(x)$ is increasing.
\end{proof}

Let $\alpha=1/{\rho}$ and use $h_1(\alpha)$ for $L_{{\rho}, u}(z)$ to
view it as a function of $\alpha$. We now compute the
derivative of $h_1(\alpha)$ w.r.t. $\alpha$:
\begin{align*}
h_1'(\alpha) &= \frac{\log[1+ \exp(\alpha (u{-}z)) ]}{ \log [1{+}
\exp(u \alpha)]} \times \\
& \quad \; \Big[\frac{\exp(\alpha (u{-}z) )} {1{+} \exp(\alpha
(u{-}z) )} \frac{u{-}z} {\log[1 {+} \exp(\alpha (u{-}z) )]} -
\frac{\exp(\alpha u )} {1 {+} \exp(\alpha u )} \frac{u} {\log[1 {+}
\exp(\alpha u )]} \Big] \\
& =\frac{\log[1+ \exp(\alpha (u{-}z)) ]}{ \alpha \log [1{+} \exp(u
\alpha)]}  \times \\
& \quad \; \Big[\frac{\exp(\alpha (u{-}z) )} {1{+} \exp(\alpha
(u{-}z) )} \frac{\log \exp(\alpha(u{-}z))} {\log[1 {+} \exp(\alpha
(u{-}z) )]} - \frac{\exp(\alpha u )} {1 {+} \exp(\alpha u )}
\frac{\log \exp(\alpha u)} {\log[1 {+} \exp(\alpha u )]} \Big].
\end{align*}
When $z<0$, we have $\exp(\alpha(u-z)) > \exp(\alpha u)$.  It then
follows from Lemma~\ref{lem:1} that $h_1'(\alpha)\geq 0$. When
$z\geq 0$, we have $h_1'(\alpha) \leq 0$ due to
$\exp(\alpha(u-z))\leq \exp(\alpha u)$. The proof of (i) is
completed.

To prove part (ii), we regard $L_{{\rho}, u}(z)$ as a
function of $u$ and denote it with $h_2(u)$. The first derivative
$h_2'(u)$ is given by
\begin{align*}
h_2'(u) & = \frac{\alpha \log[1+ \exp(\alpha (u{-}z)) ]}{ \log [1{+}
\exp(u \alpha)]} \times \\
& \quad \; \Big[\frac{\exp(\alpha (u{-}z) )} {1{+} \exp(\alpha
(u{-}z) )} \frac{1} {\log[1 {+} \exp(\alpha (u{-}z) )]} -
\frac{\exp(\alpha u )} {1 {+} \exp(\alpha u )} \frac{1} {\log[1 {+}
\exp(\alpha u )]} \Big].
\end{align*}
Using Lemma~\ref{lem:1}, we immediately obtain part (ii).

%%%%%%%%%%%%%%%%%%%%%%%%%%%%%%%%%%%%%%%%%%%%%%%%%%%%%%%%%%%%%%%%%%%%%%%%%
\section{The Proof of Theorem~\ref{thm:cohen}}

We write the objective function as
\begin{align*}
L(f) &= V_{{\rho}, u}(f) \eta + V_{{\rho}, u}(-f) (1-\eta) \\
& = {\rho} \log\big[ 1{+} \exp \frac{u {-}
f} {{\rho}} \big] \eta+ {\rho} \log\big[ 1{+} \exp \frac{u {+}
f} {{\rho}} \big] (1-\eta).
\end{align*}
The first-order and second-order derivatives of $L$ w.r.t. $f$ are given by
\begin{align*}
\frac{d L}{d f} &= - \eta \frac{\exp \frac{u {-}
f} {{\rho}}}{1{+} \exp \frac{u {-}
f} {{\rho}}} + (1-\eta) \frac{\exp \frac{u {+}
f} {{\rho}}}{ 1{+} \exp \frac{u {+}
f} {{\rho}} }, \\
\frac{d^2 L}{d f^2} &= \frac{\eta}{{\rho}}  \frac{\exp \frac{u {-}
f} {{\rho}}}{1{+} \exp \frac{u {-}
f} {{\rho}}} \frac{1}{1{+} \exp \frac{u {-}
f} {{\rho}}} + \frac{1-\eta}{{\rho}}  \frac{\exp \frac{u {+}
f} {{\rho}}}{ 1{+} \exp \frac{u {+}
f} {{\rho}} } \frac{1}{1{+} \exp \frac{u {+}
f} {{\rho}} }.
\end{align*}
Since $\frac{d^2 L}{d f^2} >0$, the minimum of $L$ is unique. Moreover, letting $\frac{d L}{d f}=0$
yields (\ref{eq:f_estimate}).

%%%%%%%%%%%%%%%%%%%%%%%%%%%%%%%%%%%%%%%%%%%%%%%%%%%%%%%%%%%%%%%%%%%%%%%%%
\section{The Proof of Proposition~\ref{thm:limit}}

%First, it is immediate that $f_{*}(\x)=0$ when $\eta=1/2$.
First, if $\eta>1/2$, we have $4\eta(1-\eta)> 4 (1-\eta)^2$ and $(2\eta-1)\exp(u/{\rho})>0$. This implies $f_{*}>0$.
When $\eta<1/2$, we have $(2\eta-1)\exp(u/{\rho})>0$. In this case, since
\[
(1{-}2\eta)^2 \exp(2u/{\rho}) + 4 \eta(1-\eta) < (1{-}2\eta)^2 \exp(2u/{\rho}) + 4 (1{-}\eta)^2 + 4 (1{-}\eta)(1{-}2\eta) \exp(u/{\rho}),
\]
we obtain $f_{*} <0$.

Second, letting $\alpha=1/{\rho}$, we express $f_{*}$ as
\begin{align*}
f_{*} &=   \frac{1} {\alpha} \log \frac{(2\eta{-}1)\exp( u \alpha) + \sqrt{(1{-}2\eta)^2 \exp(2u \alpha) + 4 \eta(1{-}\eta)} }{2 (1-\eta)}\\
       &= \frac{1}{\alpha}   \log \frac{ \frac{ (2\eta{-}1)}{|2\eta{-}1|}  + \sqrt{ 1 + \frac{4 \eta(1{-}\eta)} { (1{-}2\eta)^2 \exp(2u \alpha) } } }
{2 (1-\eta) \exp( - u \alpha)/|2\eta{-}1|}   \\
      & = \frac{1}{\alpha} \log\left[ \frac{ (2\eta{-}1)}{|2\eta{-}1|}  + \sqrt{ 1 + \frac{4 \eta(1{-}\eta)} { (1{-}2\eta)^2 \exp(2u \alpha) } } \right] -
 \frac{1}{\alpha} \log\Big[ \frac{2 (1-\eta)} {|2\eta{-}1|} \Big] + u.
\end{align*}
Thus, if $\eta>1/2$, it is clear that $\lim_{\alpha\rightarrow \infty} f_{*}=u$.  In the case that $\eta<1/2$, we have
\begin{align*}
\lim_{\alpha\rightarrow \infty} f_{*} &=  u - u \lim_{\alpha\rightarrow \infty} \frac{1 }
{-1  + \sqrt{ 1 + \frac{4 \eta(1{-}\eta)} { (1{-}2\eta)^2 \exp(2u \alpha) } } } \frac{1}{\sqrt{ 1 + \frac{4 \eta(1{-}\eta)} { (1{-}2\eta)^2 \exp(2u \alpha) } }}
\frac{4 \eta(1{-}\eta)} { (1{-}2\eta)^2 }  \exp(-2u \alpha) \\
& = u - \frac{4 \eta(1{-}\eta) u} { (1{-}2\eta)^2 } \lim_{\alpha\rightarrow \infty} \frac{  \exp(-2u \alpha)}
{-1  + \sqrt{ 1 + \frac{4 \eta(1{-}\eta)} { (1{-}2\eta)^2 \exp(2u \alpha) } } }
 \\
 & = u -  2 u \lim_{\alpha\rightarrow \infty}  {\sqrt{ 1 + \frac{4 \eta(1{-}\eta)} { (1{-}2\eta)^2 \exp(2u \alpha) } }}\\
 &=-u.
\end{align*}
Here we use  l'H\^{o}pital's rule in calculating limits.

Third, let $\alpha = \exp(u/{\rho})$. It is then immediately calculated that
\[
f_{*}'(\eta) = 2 {\rho} \frac{\alpha+ \frac{(1{-}2\eta)(1{-}\alpha^2)}{\sqrt{(1{-}2\eta)^2 \alpha^2
+ 4 \eta (1{-}\eta)}}}{ (2\eta{-}1) \alpha + \sqrt{(1{-}2\eta)^2 \alpha^2 + 4 \eta (1{-}\eta)}} + \frac{{\rho}}{1-\eta}.
\]
Consider that
\begin{align*}
A &=   \frac{2\alpha+ \frac{2(1{-}2\eta)(1{-}\alpha^2)}{\sqrt{(1{-}2\eta)^2 \alpha^2
+ 4 \eta (1{-}\eta)}}}{ (2\eta{-}1) \alpha + \sqrt{(1{-}2\eta)^2 \alpha^2 + 4 \eta (1{-}\eta)}} - \frac{1}{\eta} \\
& = \frac{\alpha-\frac{2 \eta + (1-2\eta) \alpha^2}{\sqrt{(1{-}2\eta)^2 \alpha^2
+ 4 \eta (1{-}\eta)}}} {\eta (2\eta{-}1) \alpha  + \eta \sqrt{(1{-}2\eta)^2 \alpha^2 + 4 \eta (1{-}\eta)}}.
\end{align*}
It suffices for $f_{*}'(\eta)  \geq \frac{{\rho}}{\eta(1-\eta)}$ to show $A\geq 0$. Note that
\[
\frac{(2 \eta + (1-2\eta) \alpha^2)^2}{(1{-}2\eta)^2 \alpha^2 + 4 \eta (1{-}\eta)} -\alpha^2
= \frac{4\eta^2(1{-}\alpha^2)}{(1{-}2\eta)^2 \alpha^2 + 4 \eta (1{-}\eta)} \leq 0
\]
due to $\alpha\geq 1$, with equality when and only when $\alpha=1$ or, equivalently, $u=0$.
Accordingly, we have $\alpha-\frac{2 \eta + (1-2\eta) \alpha^2}{\sqrt{(1{-}2\eta)^2 \alpha^2
+ 4 \eta (1{-}\eta)}} \geq 0$.

%%%%%%%%%%%%%%%%%%%%%%%%%%%%%%%%%%%%%%%%%%%%%%%%%%%%%%%%%%%%%%%%%%%%%%%%%
\section{The Proof of Theorem~\ref{thm:consistent2}}

In order to prove the theorem,  we define
\[
\delta_{\gamma} := \sup\{t: \gamma t^2 \leq 2 V_{{\rho}}(0) \} = \sqrt{2/\gamma}
\]
for $\gamma >0$ and let $V_{{\rho}}^{(\gamma)}(y f) $ be the coherence function $V_{{\rho}}(y f)$ restricted to $\YM {\times} [-\delta_{\gamma}k_{\max},  \delta_{\gamma}k_{\max}]$,
where $k_{\max}= \max_{\x \in \XM} K(\x, \x)$. For the Gaussian RBF kernel, we have $k_{\max}=1$.

It is clear that
\[
\|V_{{\rho}}^{(\gamma)}\|_{\infty} := \sup\big\{V_{{\rho}}^{(\gamma)}(y f), (y, f) \in  \YM {\times} [-\delta_{\gamma}k_{\max},  \delta_{\gamma}k_{\max}] \big\}
= {\rho} \log\Big(1 + \exp\frac{u {+} k_{\max}\sqrt{2/\gamma}}{{\rho}}\Big).
\]
Considering that
\[
\lim_{\gamma \rightarrow 0} \frac{ \|V_{{\rho}}^{(\gamma)}\|_{\infty}}{k_{\max} \sqrt{2/\gamma}} =
\lim_{\alpha \rightarrow \infty} \frac{\exp \frac{u {+} \alpha}{{\rho}}} { 1 + \exp \frac{u {+} \alpha}{{\rho}}} =1,
\]
we have $\lim_{\gamma \rightarrow 0} \|V_{{\rho}}^{(\gamma)}\|_{\infty}/ \sqrt{1/\gamma} =  \sqrt{2} k_{\max}$. Hence,
we have $\|V_{{\rho}}^{(\gamma)}\|_{\infty} \sim  \sqrt{1/\gamma}$.

On the other hand, since
\[
V_{{\rho}}^{(\gamma)}(y f) = V_{{\rho}}^{(\gamma)}(y f_1) - \frac{\partial V_{{\rho}}^{(\gamma)}(y f_2)} {\partial f} (f -f_1),
\]
where $f_2 \in [f, f_1] \subseteq [-\delta_{\gamma}k_{\max},  \delta_{\gamma}k_{\max}]$,
we have
\begin{align*}
|V_{{\rho}}^{(\gamma)}|_{1} &:= \sup\left\{ \frac{|V_{{\rho}}^{(\gamma)}(y f){-} V_{{\rho}}^{(\gamma)|}(y f_1)}{|f-f_1|}, \; y \in  \YM,
f, f_1 \in [-\delta_{\gamma}k_{\max},  \delta_{\gamma}k_{\max}], f\neq f_1 \right\} \\
& = \sup\left\{ \Big| \frac{\partial V_{{\rho}}^{(\gamma)}(y f_2)} {\partial f} \Big|, \; y \in  \YM, f_2 \in [-\delta_{\gamma}k_{\max},  \delta_{\gamma}k_{\max}] \right\} \\
& = \frac{\exp\frac{u {+} k_{\max}\sqrt{2/\gamma}}{{\rho}}} {1 + \exp\frac{u {+} k_{\max}\sqrt{2/\gamma}}{{\rho}}}.
\end{align*}
In this case, we have $\lim_{\gamma \rightarrow 0} |V_{{\rho}}^{(\gamma)}|_{1} =1$, which implies that $|V_{{\rho}}^{(\gamma)}|_{1} \sim 1$.

We now immediately conclude Theorem~\ref{thm:consistent2} from Corollary~3.19 and Theorem~3.20 of \citet{SteinwartTIT:2005}.

%%%%%%%%%%%%%%%%%%%%%%%%%%%%%%%%%%%%%%%%%%%%%%%%%%%%%%%%%%%%%%%%%%%%%%%%%%%%%%%%%%%%%%%%%%%%%%%%%
%%%%%%%%%%%%%%%%%%%%%%%%%%%%%%%%%%%%%%%%%%%%%%%%%%%%%%%%%%%%%%%%%%%%%%%%%%%%%%%%%%%%%%

%%%%%%%%%%%%%%%%%%%%%%%%%%%%%%%%%%%%%%%%%%%%%%%%%%%%%%%%%%%%%%%%%%%%%%
%%%%%%%%%%%%%%%%%%%%%%%%%%%%%%%%%%%%%%%%%%%%%%%%%%%%%%%%%%%%%%%%%%%%%%%%
%\small
%\bibliographystyle{plain}
\bibliography{clearning2}

\end{document}